\documentclass[10pt,journal]{IEEEtran}

\usepackage{graphicx}

\usepackage{amsmath}
\usepackage{amssymb}
\usepackage{mathtools, nccmath}

\usepackage{subcaption}

\usepackage[table,xcdraw]{xcolor}

\usepackage{comment}

\newif\iftrackchanges
\trackchangestrue

\renewcommand{\vec}[1]{\boldsymbol{#1}}

\newcommand{\leftimage}{I_{L}}
\newcommand{\rightimage}{I_{R}}
\newcommand{\costvolume}{V}
\newcommand{\truncatedcostvolume}{\costvolume^t}
\newcommand{\truncatedcostvolumesupport}{\chi}
\newcommand{\bnnInputs}{\vec{x}}

\newcommand{\truedisparity}{d}			                    
\newcommand{\rawdisparity}{\truedisparity_{raw}}		
\newcommand{\therefineddisparity}{\hat{\truedisparity}}

\newcommand{\disprefiner}{\delta}
\newcommand{\estimatordisprefiner}{\hat{\disprefiner}}            

\newcommand{\stereoalgo}{S}					
\newcommand{\trainingset}{D}

\newcommand{\nn}{\Phi}                       

\newcommand{\bnnparams}{\vec{\theta}} 
\newcommand{\variationalparams}{\vec{\phi}} 

\newcommand{\ie}{i.e., }
\newcommand{\etal}{\textit{et al.}}

\newcommand{\ourdatasetname}{Active-Passive SimStereo}

\begin{document}

\title{Bayesian Learning for Disparity Map Refinement for Semi-Dense Active Stereo Vision}

\author{Laurent Valentin Jospin,
        Hamid Laga,
        Farid Boussaid,
        and~Mohammed Bennamoun,~\IEEEmembership{Senior Member,~IEEE}
\thanks{L.V. Jospin, F. Boussaid and M. Bennamoun are with the University of Western Australia.}
\thanks{H. Laga is with Murdoch University and the University of South Australia.}
}

\maketitle

\begin{abstract}
A major focus of recent developments in stereo vision has been on how to obtain accurate dense disparity maps in passive stereo vision. Active vision systems enable more accurate estimations of dense disparity compared to passive stereo. However, subpixel-accurate disparity estimation remains an open problem that has received little attention. In this paper, we propose a new learning strategy to train neural networks to estimate high-quality subpixel disparity maps for semi-dense active stereo vision. The key insight is that neural networks can double their accuracy if they are able to jointly learn how to refine the disparity map while invalidating the pixels where there is insufficient information to correct the disparity estimate. Our approach is based on Bayesian modeling where validated and invalidated pixels are defined by their stochastic properties, allowing the model to learn how to choose by itself which pixels are worth its attention. Using active stereo datasets such as \ourdatasetname{}, we demonstrate that the proposed method outperforms the current state-of-the-art active stereo models. We also demonstrate that the proposed approach compares favorably with state-of-the-art passive stereo models on the Middlebury dataset.
\end{abstract}


\section{Introduction}

\IEEEPARstart{T}{he} past few years have seen a surge in the use of deep learning networks for disparity estimation in stereo vision~\cite{laga2020survey,9395220}. State-of-the-art methods use an end-to-end approach where a network is trained for all the steps of the stereo matching pipeline, including feature extraction, cost volume construction and regularization, as well as disparity map refinement. Most of these methods, which use deep neural networks with large and multiscale receptive fields,  have been designed to address challenges faced by passive stereo systems such as how to estimate accurate disparities up to the pixel level in textureless regions and in regions with repetitive patterns and self-similarities. Compared to passive stereo, active stereo-based techniques use projected patterns to facilitate the matching, and thus the disparity computation, by removing textureless areas and regions with self-similar patterns from the image~\cite{Keselman_2017_CVPR_Workshops}.

\begin{figure}[t]
    \centering
    \includegraphics[width=0.45\textwidth]{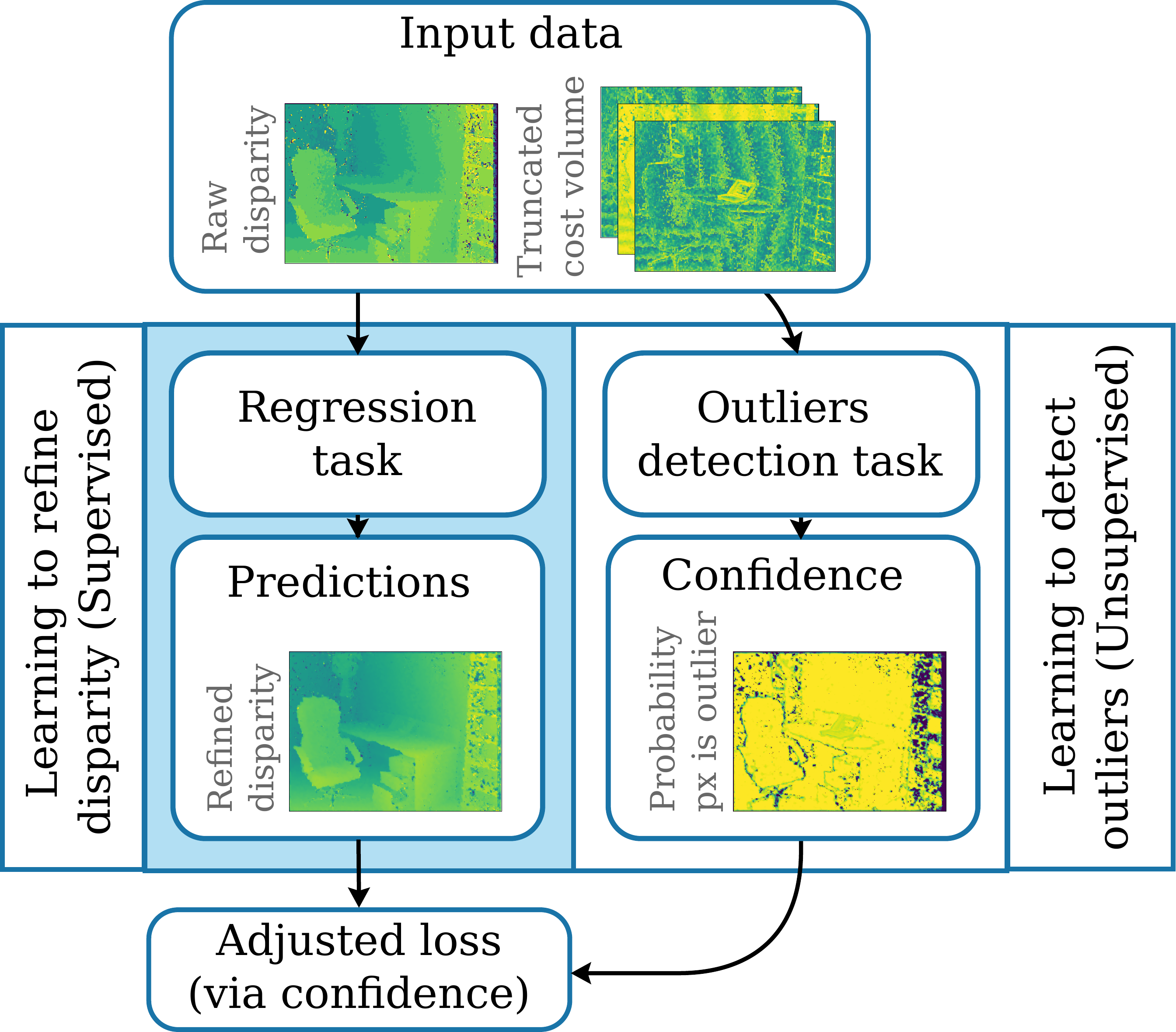}
    \caption{The proposed disparity refinement approach uses the model confidence, learned in an unsupervised manner, to adjust the learning behavior (via a confidence-dependent loss) and help the network learn which image areas are worth focusing on in order to increase its precision.}
    \label{fig:problem}
\end{figure}

Motivated by the success of end-to-end models, several papers applied to active stereo technics that have been originally developed for passive stereo~\cite{Zhang_2018_ECCV,9360804}. While these borrowed architectures were optimized to guarantee a dense reconstruction of the disparity maps, which is a major problem in passive stereo, they are not meant to address challenges such as accurate subpixel disparity refinement. This makes such models redundant in the case of active stereo where simpler methods will correctly match a large proportion of pixels without requiring large and expensive modules. However, subpixel accuracy is still an open issue. We argue in this paper that active stereo vision should be tackled with a different approach, centered around subpixel refinement. The proposed model should be lightweight to enable its use for embedded platforms~\cite{puglia2020deep,Shi2021} such as smartphones, virtual and augmented reality headsets,  and consumer-grade stereo cameras~\cite{Keselman_2017_CVPR_Workshops}.

Since active stereo vision ensures a high proportion of pixels correctly matched even with low-end matching algorithms, it makes \emph{semi-dense stereo matching} an interesting approach. Semi-dense stereo matching~\cite{semi-dense-stereo2019}, which is the process of producing a disparity map where some pixels, considered as outliers, are discarded and not used in the subsequent steps of the process. This is especially relevant for applications such as Augmented Reality (AR) \cite{Orts-EscolanoSergio2016Holoportation} where the visual quality of the depth map is important, but some pixels can be discarded without impacting the performances of downstream tasks. This is critical for lightweight models, which cannot hold as much information as larger scale ones \cite{7133169,thompson2020computational}. This derives naturally from the Data processing inequality \cite{9468892}. We discuss these issues in more detail in the supplementary material, where we provide theoretical guarantees on our semi-dense model benefits.

In this paper, we propose a novel method that is optimized for semi-dense active stereo. The method jointly learns both aspects of semi-dense stereo matching, \ie identifying outliers and refining raw disparity maps at the sub-pixel accuracy level. We first propose to characterize the outliers via a Bayesian probabilistic model where the only meta-parameter to set is the desired reconstruction accuracy. The benefit of using a Bayesian model is that it allows us to account not only for aleatoric uncertainty (\ie the baseline uncertainty present in the data) but also for epistemic uncertainty (\ie the uncertainty due to the lack of information when training the model)~\cite{Kendall2017WhatVision}. We then propose a novel deep neural network architecture (Fig.~\ref{fig:problem}) composed of an outlier detection branch, trained in an unsupervised manner, and a refinement branch trained in a supervised manner. The role of the outlier detection branch is to invalidate, during training, the pixels that are outliers so that the refinement network focuses on the inliers hereinafter referred to as \emph{validated} pixels, which are easier to reconstruct. The outlier detection branch can be seen as a form of meta learning that influences the loss derived from a Bayesian Probabilistic Graphical Model (PGM) and used to train the refinement branch. At test time, the outlier detection branch can still be used to clean up the disparities predicted by the model.

We demonstrate, in the context of semi-dense active stereo vision, that the proposed lightweight Bayesian Neural Network (BNN) based on this principle is able to double the accuracy of the disparity refinement at the validated pixels compared to a similar refinement network trained without the detection and invalidation of the outliers. Despite its small size, the proposed architecture achieves state-of-the-art performance, with errors of the order of $0.2$~pixel in the validated areas. 
We also show that the proposed BNN trained on active stereo is able to generalize to passive stereo, albeit at the cost of a drastically reduced proportion of inliers. This demonstrates that the proposed architecture allows the network to be aware of when it can safely make a prediction, a very desirable property of BNNs. Finally, the proposed architecture is significantly faster than the state-of-the-art both at training and test time. The main contributions of this paper can be summarized as follows:

\begin{itemize}
    \item We propose a new approach for fast context aggregation to allow networks to operate with a limited number of hidden layers (Sec.~\ref{sec:ourapproach:functional:ctx}).
    
    \item We propose a novel unsupervised Bayesian approach to learn invalidation masks that enable the refinement process to focus on image areas that are likely to lead to improved accuracy (Sec.~\ref{sec:ourapproach}). The proposed invalidation mechanism is robust to domain shifts since the BNN only identifies as inliers those regions for which it can make accurate predictions. Moreover, despite a small model size, the target subpixel precision can still be achieved, as no information needs to be stored for outliers. As a result, the method is easier to deploy on embedded platforms compared to state-of-the-art deep learning models. To the best of our knowledge, this is the first time BNNs have been used not only to predict uncertainty but also to improve the accuracy of the model predictions.
    
    \item Our experiments (Sec.~\ref{sec:evaluation}) demonstrate that our novel BNN-based approach significantly increases the precision of disparity map refinement for semi-dense active stereo matching problems.
    
\end{itemize}

\noindent The remainder of the paper is organized as follows; Section~\ref{sec:related_work} reviews the related work. Section~\ref{sec:ourapproach} details the proposed model and training approach. Section~\ref{sec:evaluation}  evaluates the performance of the proposed model and discusses its generalization ability under domain shift. Section~\ref{sec::conclusion} concludes the paper.


\section{Related work}
\label{sec:related_work}

Stereo matching algorithms usually address one or more of the following three challenges: \textbf{(1)} disparity map densification, \ie how to increase the number of patches that are correctly matched across a pair of images. Usually, a match is considered correct if the matching error is within two or three pixels.  \textbf{(2)} Outliers removal, which is the problem of detecting and removing incorrectly matched pixels from the estimated disparity map.  \textbf{(3)} Disparity map refinement, which aims to improve the resolution and accuracy of the disparity, and thus depth, maps. This paper focuses on the second and third problems. The first one has been extensively explored; see the surveys of Han \etal~\cite{HanXian-Feng2019Image-basedEra}, Laga \etal~\cite{laga2020survey}, and Poggi \etal~\cite{9395220} for more details.

\subsection{Real time stereo models}

A major issue of the state-of-the-art stereo models is their computational cost \cite{Chang_2020_ACCV}. Powerful and expensive GPUs are often required to achieve an interactive frame rate. Different mitigation strategies have been proposed: For example, Wang \etal{} proposed AnyNet \cite{AnyNet2019}, a hierarchical model where successive levels of disparities are refined in succession from low to high resolution. CascadeStereo \cite{gu2019cas} was proposed to leverage a coarse to fine approach making end-to-end matching networks more memory efficient and thus better suited to handle high-resolution images. RealTimeStereo \cite{Chang_2020_ACCV} introduced an attention-aware feature aggregation module to reduce the dimension of the feature space while using a hierarchical approach similar to AnyNet. MobileStereoNet \cite{shamsafar2022mobilestereonet}, on the other hand, leverage the optimized activation functions of MobileNet \cite{howard2019searching} for stereo matching. To reduce the computation time, Yee and Chakrabarti \cite{Yee_2020_WACV} proposed to remove the initial feature computation module and instead compute the cost volume using traditional methods. Their method uses a Neural network only to infer the disparity from the cost volume. Rahim \etal{} \cite{SeparableConvolutionsStereo} also proposed to use separable convolutions to speed up the processing of the cost volume.

All the aforementioned methods trade accuracy to gain speed. In contrast, because our approach is based on semi-dense stereo, it sacrifices the density of the depth map instead of accuracy.

\subsection{Outliers detection and semi-dense stereo matching}

The aim of outlier detection is to ensure that a subsequent process is not adversarially affected by a small number of out-of-distribution samples~\cite{doi:10.1002/widm.2}. This is very important for applications such as augmented reality and digital art, which require accurate 3D reconstruction.  Most outlier detection methods in stereo matching use cues in the cost volume~\cite{Poggi2017EfficientStereo}, but some methods also use the estimated raw disparities as a source of information \cite{semi-dense-stereo2019}.

When used for invalidation~\cite{Zhang_2018_ECCV,doi:10.1111/cgf.13753} (\ie for discarding pixels whose error is high), the outlier detection task operates jointly with another task. The latter is referred to as the main task. Since outliers are defined by their impact on the main task, one efficient learning strategy is to jointly learn how to detect them and perform the main task~\cite{Yi_2018_CVPR}.  Bevandi{\'{c}} \etal~\cite{10.1007/978-3-030-33676-9_3} showed that outlier detection in a multi-task setting can share features with semantic segmentation without degrading the performance of the main or outlier detection tasks.  Xu \etal~\cite{xu2006robust} showed that detecting and invalidating outliers during training can improve training performances. However, because their approach solely attaches a confidence variable to each training sample, it cannot detect outliers at runtime. In contrast, our proposed approach improves the training performance while being able to detect outliers at runtime.

The process of invalidating outliers in stereo vision is called semi-dense stereo vision. This has been extensively investigated using traditional methods~\cite{Hu2012AVision} and more recently using deep learning techniques~\cite{Poggi2017QuantitativeWorld}. Whether a pixel is an outlier depends on the technique used to generate the disparity. This implies that no labeled data can be prepared before the training is actually conducted so that a network can learn how to detect its own outliers. Consequently, the training of an outlier detection network has to be performed in an unsupervised or a self-supervised manner by using auxiliary supervisory signals such as left-right or multi-view consistency~\cite{Jie2018Left-RightMatching,Zhang_2018_ECCV,MostegelContradict2016}. Outliers can also be defined using the regression module's prediction errors~\cite{Shaked_2017_CVPR,LAFNet2019,semi-dense-stereo2019}. In our case, a Bayesian model is used to define outliers based on their expected stochastic properties. This provides multiple advantages over the aforementioned methods. The most important one is that our model accounts
for not only aleatoric uncertainty but also epistemic uncertainty. As a result, our model is less likely to fail to recognize error cases it did not see at training time.

\subsection{Disparity refinement}

Various traditional methods have been proposed to deal with noise in stereo matching. Examples include smoothing either the estimated raw disparities or the cost volumes using low-pass filters~\cite{Vetterli2014}, morphological operators,  simple regularizers~\cite{RUDIN1992259},  or adaptive filtering~\cite{Wu2014realTimeshaderefinedepth, Zhang16denoising} guided either by an RGB frame, if available, or by the raw depth map itself~\cite{Yan2018DDRNet:CNNs, Batsos2018RecResNet:Enhancement}.  Other methods rely on interpolation \cite{jospin2021generalized} either in the input image space~\cite{4409212,6460147} or in the cost volume space \cite{Shimizu2005}. Interpolation in the cost volume is computationally cheaper but does not have a closed-form solution that is both optimal and universal since the optimal interpolation function depends on the cost function used for matching~\cite{Miclea2015}. In the latest end-to-end deep learning approaches for stereo matching, the cost volume is generally interpolated on the fly via the use of the softargmax or softargmin operators~\cite{laga2020survey}, which offers the benefits of giving a smooth estimate of the disparity from the cost volume in a differentiable fashion. The use of the softargmax or softargmin operators is a key feature for end-to-end models but offer poor performances in term of subpixel accuracy. Our approach is based on cost-volume interpolation, which significantly reduces the memory and computation footprint required. Since we are using a learning based method, the lack of a universal interpolation formula~\cite{Miclea2015} is not an issue, since the proposed network learns an optimal solution for the matching cost at hand directly from the data.

\subsection{Active stereo vision}

In contrast to passive stereo, active vision has received little attention from the deep learning community. One major issue which hampered the development of active stereo method was the lack of good quality datasets. Yet, a few self-supervised models have been proposed such as ActiveStereoNet \cite{Zhang_2018_ECCV} and TSFE-Net \cite{9360804}. ActiveStereoNet~\cite{Zhang_2018_ECCV} is based on the architecture of StereoNet~\cite{Khamis_2018_ECCV}, an end-to-end deep learning model, with an edge-aware upsampling module, originally developed for passive stereo setups. The main contribution of ActiveStereoNet is not the model itself but its self-supervised training method, which was developed to overcome the lack of active stereo datasets suitable for training. Another end-to-end architecture was employed by TSFE-Net~\cite{9360804}.  Utilizing a local contrast normalization module, the authors demonstrate how the network can exploit the relationship between the speckle intensity and the distance from the active pattern projector.  Unlike these models, which mimic the architectures used for passive stereo, our approach focuses on the refinement step in the stereo pipeline. Using our active stereo model design, one can achieve better subpixel accuracy at a fraction of the computational cost of conventional approaches (Sec.~\ref{sec:evaluation:comp}).

\begin{figure}[t]
    \centering
    \includegraphics[width=0.45\textwidth]{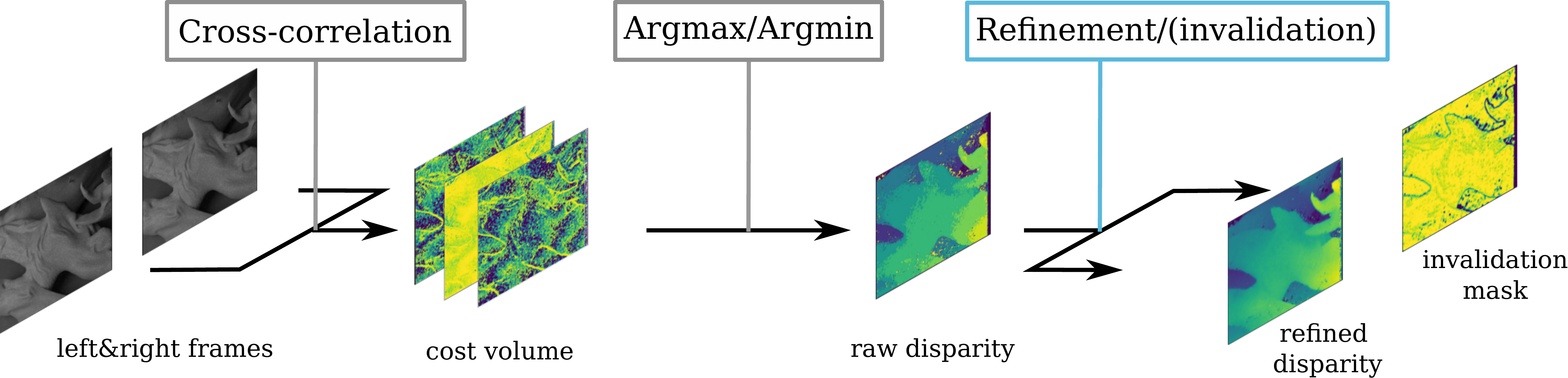}
    \vspace{-5pt}
    \caption{Standard semi-dense stereo matching pipeline. This paper contribution relates to the last step (see Fig.~3). }
    \label{fig:stereopipeline}
\end{figure}

\section{Proposed Approach}
\label{sec:ourapproach}

The focus of this paper is on the refinement steps of the stereo matching pipeline. Hence, to build the initial disparity estimate and cost volume, we use a standard coarse to fine hierarchical algorithm \cite{faugeras1993real} based on the Zero-Mean Normalized correlation (ZNCC) matching cost \cite{Hirschmuller2009EvaluationDifferences}. The ZNCC function was initially proposed to match images with different bias and gain factors. We choose this function as it ensures the cost is normalized with respect to the input image, which should increase the robustness of our refinement model to variations in the input images intensities. This coarse to fine hierarchical algorithm is significantly faster than the current state-of-the-art deep learning-based cost volume construction and regularization, making it ideal for lightweight systems. The implementation we are using, which runs on the CPU, requires around $60$ms per frame with $160$ disparity levels. This is negligible when compared to more than $2$ seconds for ActiveStereoNet running on the CPU, or $500$ms for our refinement (see Table~\ref{table:modelfoootprint}). 

To perform the refinement and outlier pixels invalidation, we propose two modules that work collaboratively: a supervised disparity refinement module and an unsupervised outlier pixel invalidation module (Fig.~\ref{fig:problem}). In contrast to previous studies, we identify the outliers corrupting a disparity map based on their stochastic properties, which we model using a Probabilistic Graphics Model (PGM), hereinafter referred to as the stochastic model (Section~\ref{sec:ourapproach:stochastic}), rather than an error threshold \cite{semi-dense-stereo2019} or a left-right consistency criterion \cite{Zhang_2018_ECCV}.

Let $\rawdisparity = \stereoalgo(\leftimage, \rightimage)$ be a raw disparity map estimated  using a stereo matching algorithm $\stereoalgo$  from a pair of left $\leftimage$ and right $\rightimage$ images. We define the refinement process as a function of the raw disparity  $\rawdisparity$ and the matching cost volume $\costvolume$. To minimize the amount of data the model has to process, thus decreasing its memory and computational footprint, we only consider a  version of the cost volume that is truncated along the disparity dimension, hereinafter denoted by $\truncatedcostvolume$, which is  defined  at pixel $k =(i, j)$ as:
\begin{equation}
    \truncatedcostvolume(k) = \left\{ \begin{array}{rl}
        \costvolume(\rawdisparity+k) & \text{if~} ~ \rawdisparity+k \in \left[0,d_{max}\right], \\
        0 & \text{otherwise.}
    \end{array} \right.
\end{equation}
\noindent Let also $\truncatedcostvolumesupport$ be a tensor defined at pixel $(i, j)$ as:
\begin{equation}
    \truncatedcostvolumesupport(k) = \left\{ \begin{array}{rl}
        1 & \text{if~} ~ \rawdisparity+k \in \left[0,d_{max}\right], \\
        0 & \text{otherwise.}
    \end{array} \right.
\end{equation}
\noindent Here, $d_{max}$ is the maximum disparity under consideration. We designate a tuple of inputs $(\rawdisparity, \truncatedcostvolume, \truncatedcostvolumesupport)$ as $\bnnInputs$. 

\begin{figure}[t]
    \centering
    \includegraphics[width=0.45\textwidth]{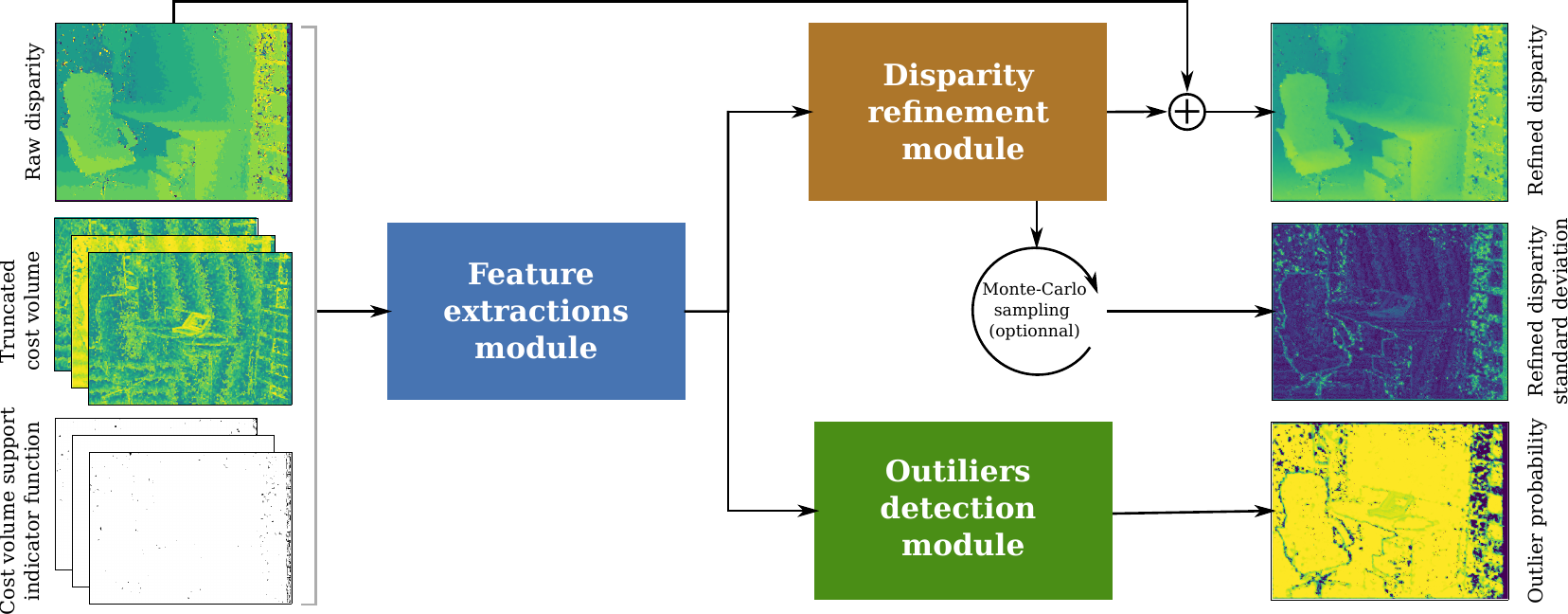}
    \caption{The general structure of the proposed module.}
    \label{fig:ourpipeline}
\end{figure} 

Let $\disprefiner$ be the error in the initial disparity map $\rawdisparity$. Our goal is to learn an estimator $\estimatordisprefiner(\bnnInputs)$ defined as a function of $\rawdisparity, \truncatedcostvolume$, and $\truncatedcostvolumesupport$. In other words, the true disparity $\truedisparity$  can be written as:
\begin{equation}
    \label{eq:dispfromraw}
   \truedisparity = \rawdisparity + \estimatordisprefiner(\bnnInputs) + \varepsilon, ~ \varepsilon \sim \mathcal{N}(0, \sigma_{\varepsilon}^2).
\end{equation}
\noindent We assume that the error between $\disprefiner$ and $\estimatordisprefiner$, denoted by $\varepsilon$, follows a normal distribution with mean $0$ and standard deviation $\sigma_{\varepsilon}$.

To learn $\hat{\disprefiner}(\bnnInputs)$, we use a BNN whose architecture, or functional model, is composed of a feature extraction, a disparity refinement, and an outlier detection module (Fig.~\ref{fig:ourpipeline}); see Section~\ref{sec:ourapproach:functional} for more details. 

Training a BNN involves finding the probability distribution $p(\bnnparams|\trainingset)$ of its parameters $\bnnparams$ knowing the training set $\trainingset$. This is difficult to achieve in practice. Instead, we use variational inference and Monte-Carlo dropout as an approximation~\cite{jospin2020handson}. To simplify the predictions at runtime,  we also limit the number of Bayesian layers that we position at the end of the network \cite{zeng2018relevance,brosse2020lastlayer}. Variational inference is the process of learning the parameters $\variationalparams$   of a  distribution $q_{\variationalparams}(\bnnparams)$ that closely approximates $p(\bnnparams|\trainingset)$. Monte-Carlo dropout is the process of using dropout layers at runtime as a form of variational inference~\cite{jospin2020handson}. The loss for learning $\variationalparams$ is derived from the stochastic model (Fig.~\ref{fig:pgm}). Since the training process is stochastic, epistemic uncertainty, \ie the uncertainty on the model itself rather than the uncertainty due to noise in the data, is properly accounted for. This is an important benefit of using a BNN.

At runtime, the prediction of a BNN and its uncertainty are usually obtained via Monte-Carlo sampling, \ie by running the network multiple times with different sets of weights sampled from the posterior. However, this is not practical for real-time applications. Since we are using Monte-Carlo dropout and variational layers positioned at the end of the network, an estimate of the mean prediction can be obtained in a single pass by flattening the dropout layers and using the mean weights for the variational layers. Using a stochastic neural network only for training seems redundant, but doing so allows the outlier detection module to be properly calibrated based on the uncertainty of the network during training. At runtime, while the marginal distribution of the refined disparity can still be estimated by running multiple passes, the uncertainty is mostly given by the outlier detection branch of the network. This preserves most of the advantages of BNNs, especially their robustness to outliers, while significantly reducing the computation time required to estimate the uncertainty. We detail each of these modules in the following subsections.

\subsection{Stochastic model}
\label{sec:ourapproach:stochastic}

\begin{figure}[t]
    \centering
    \includegraphics[width=0.3\textwidth]{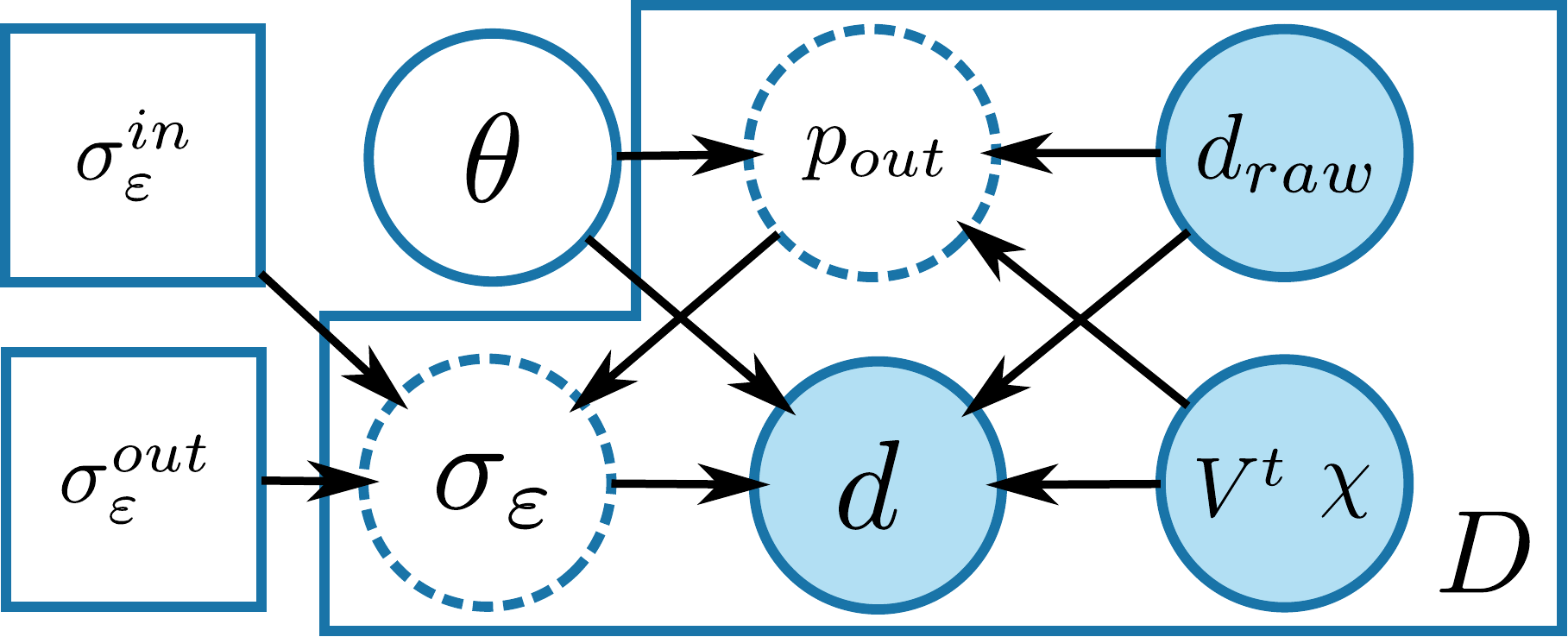}
    \vspace{-3pt}
    \caption{The probabilistic graphical model corresponding to our approach.}
    \label{fig:pgm}
\end{figure}

\begin{figure*}[t]
    \centering
    \includegraphics[width=0.95\textwidth]{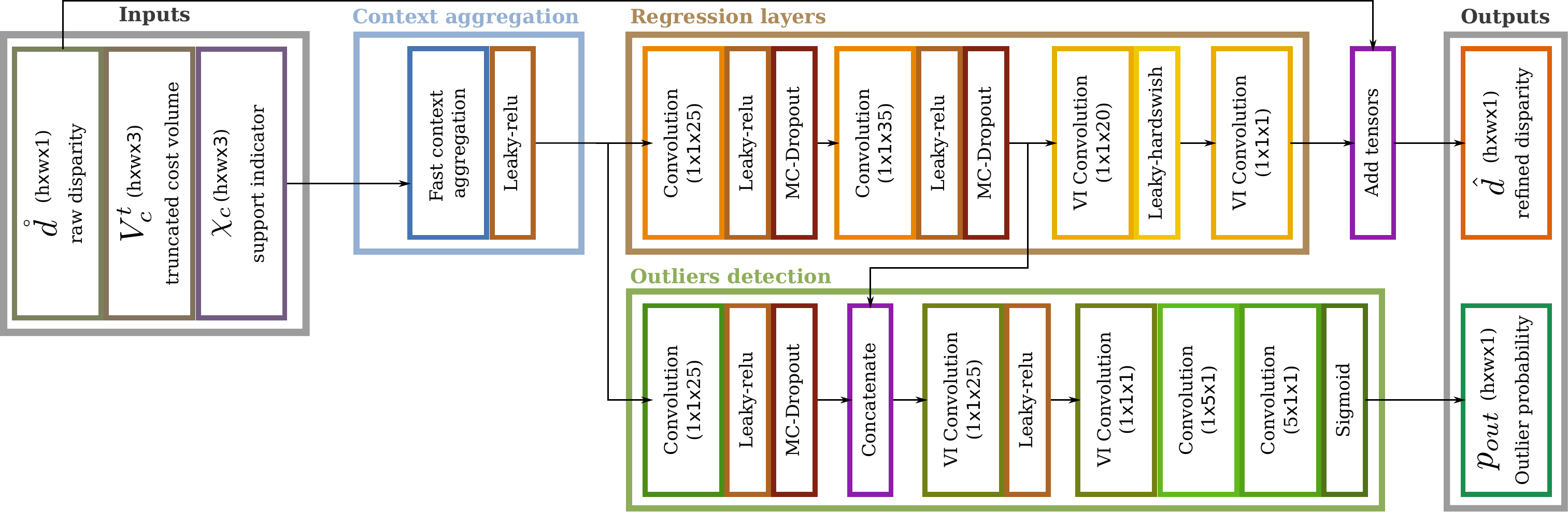}
    \caption{Architecture of the proposed Convolutional Neural Network architecture. The point estimate convolution layers are designated as \emph{Convolution}, while the variational convolution layers are designated as \emph{VI convolution}.}
    
    \label{fig:archi}
\end{figure*}

We use a PGM (Fig.~\ref{fig:pgm}) to describe the stochastic properties of the considered model. This means that we have to specify $p(d |o, \bnnInputs, \bnnparams)$, $p(o |\bnnInputs, \bnnparams)$, and $p(\bnnparams)$, where $o$ is a binary stochastic variable indicating whether the pixel under consideration is an outlier or not. Here, $p(\bnnparams)$ is simple to define as the BNN parameters do not depend on any other variables. We chose a Gaussian prior with a standard deviation $\sigma_{\bnnparams}=1$. The choice of $\sigma_{\bnnparams}$ sets the amount of $\ell_2$ regularization for the network weights $\bnnparams$ \cite{jospin2020handson}. It has been chosen based on the expected range of corrections to apply, which should be around one pixel on average.

The conditional probability $p(o | \bnnInputs, \bnnparams)$ is a Bernoulli distribution. We do not set its probability to a constant value in the prior. Instead, we assume it is a function of $\bnnInputs$ and $\bnnparams$, which is given by the functional model; see Section~\ref{sec:ourapproach:functional}. As long as the prior $p(\bnnparams)$ ensures that all values between $0$ and $1$ are equiprobable for the probability of a given pixel being an outlier, our model makes no assumptions about the proportion of outliers in the training data. Instead, it learns the appearance of an outlier directly from the data. This can be understood as being a type of hierarchical Bayes, a formulation used to perform Bayesian meta-learning \cite{hospedales2020metalearning}. In the final model, we marginalize the variable $o$ since learning to predict the probability $p_{out}$ that a given pixel is an outlier is already informative enough in itself:
\begin{equation}
    \label{eq:omarginalized}
    p(d |\bnnInputs, \bnnparams) = \int p(d |o, \bnnInputs, \bnnparams) p(o |\bnnInputs, \bnnparams) do .
\end{equation}

\begin{table*}[t]
    \centering
    
    \scalebox{1.0}{
    \begin{tabular}{|ll|r|r|r|r|r|r|}
    \hline
        \multicolumn{2}{|l|}{} & w/ $p_{out}$ &  w/o $p_{out}$ & ActiveStereoNet \cite{Zhang_2018_ECCV} & MobileStereoNet \cite{shamsafar2022mobilestereonet} & ACVNet \cite{xu2022attention} & CascadeStereo \cite{gu2019cas}\\
    \hline
        \multicolumn{8}{|l|}{Parameters}  \\
    \hline
      & Base network & $19,180$ & $12,824$ & $447,492$ & $2,352,706$ & $7,173,488$ & $10,943,776$\\
      & Additional Variational & $5,081$ & $941$ & $0$ & $0$ & $0$ & $0$\\
      & \textbf{total} & $24,261$ & $13,765$ & $447,492$ & $2,352,706$ & $7,173,488$ & $10,943,776$ \\
    \hline
        \multicolumn{8}{|l|}{Memory footprint [MB] for images of size $640\times480$} \\
    \hline
      & Model & $0.097$ & $0.055$ & $1.790$ & $9.411$ & $28.694$ & $43.775$ \\
      & Tensors (w/ gradient) & $756.361$ & $462.274$ & $2,106.368$ & $20,436.285$ & $13,230.648$ & $13,802.343$\\
      & Tensors (w/o gradient) & $11.148$ & $9.915$ & $421.904$ & $4,229.265$ & $4,216.855$ & $4,225.285$ \\
    \hline
        \multicolumn{8}{|l|}{One pass time [ms] for images of size $640\times480$}  \\
    \hline
      & GPU {\scriptsize (GeForce RTX 2080 Ti)} & ~9.75\textsuperscript{1} & ~7.17\textsuperscript{1} & $22.22$ to $53.59$ \textsuperscript{2} & $112$ to $119$ \textsuperscript{2} & $331$ \textsuperscript{3} & $234$ \textsuperscript{3}\\
      & CPU {\scriptsize (20 x i9-10900X @ 3.70GHz)~} & 471\textsuperscript{1} & 398\textsuperscript{1} & ~$1,152$ to $2,783$  \textsuperscript{2} & ~$4,182$ to $5,186$ \textsuperscript{2}  & ~$13,254$  \textsuperscript{3} & $8,017$ \textsuperscript{3}\\
    \hline
    \end{tabular}}
    \caption*{\tiny \textsuperscript{1} Our model also requires an additional 45 to 83 \textsuperscript{2} ms on cpu for the initial matching, but compared to the cpu time of either our model or ActiveStereoNet, this is negligible. \\ 
    \textsuperscript{2} The computation time varies depending on the maximum disparity. Here, computed for the range $20$ to $160$ px. \\ 
    \textsuperscript{3} Disparity range has to be $192$px due to the network architecture.}
    \caption{Comparison of our network architecture and its footprint with the state-of-the-art.}
    \label{table:modelfoootprint}
    
\end{table*}

\noindent According to Equation~\eqref{eq:dispfromraw} the conditional probability $p(d |o, \bnnInputs, \bnnparams)$ is set as a Normal distribution with mean $\rawdisparity + \estimatordisprefiner(\bnnInputs)$ and standard deviation $\sigma_\varepsilon$. We assume $\sigma_\varepsilon$ is equal to $\sigma_{\varepsilon}^{in}$  for inliers and $\sigma_{\varepsilon}^{out}$ for outliers. Since $o$ is marginalized, we can approximate the solution to Equation~\eqref{eq:omarginalized} by a normal distribution of mean $\rawdisparity + \estimatordisprefiner(\bnnInputs)$ and whose standard deviation can be approximated by combining $\sigma_{\varepsilon}^{in}$  and $\sigma_{\varepsilon}^{out}$:
\begin{equation}
    \label{eq:modelLikelyhood}
    p(d |\bnnInputs, \bnnparams) \sim \mathcal{N}(\rawdisparity + \estimatordisprefiner(\bnnInputs), (\sigma_{\varepsilon}^{in} + (\sigma_{\varepsilon}^{out} - \sigma_{\varepsilon} ^{in})p_{out})^2).
\end{equation}
The posterior given the training set $\trainingset$, up to a scaling constant, is then:
\begin{equation}
    p(\bnnparams | \bnnInputs, \trainingset) \propto \left( \prod_{d \in \trainingset} p(d |\bnnInputs, \bnnparams) \right) p(\bnnparams).
\end{equation}
\noindent This formulation can be seen as the dual of learning while dealing with noisy labels, a problem related to ours that has been explored in the literature~\cite{6685834}. This includes meta learning approaches where a stochastic model is learned to adjust the loss to account for the noise in the training set~\cite{Wang_2020_CVPR,Patrini_2017_CVPR}. The proposed stochastic model deals with the uncertainty of the BNN, instead of dealing with the noise in the training set.

\subsection{Functional model}
\label{sec:ourapproach:functional}

We use a single Bayesian CNN as an estimator for both the refined disparity $\therefineddisparity$ and the probability map $p_{out}$:
\begin{equation}
    (\therefineddisparity, p_{out}) = \nn(\rawdisparity, \truncatedcostvolume, \truncatedcostvolumesupport).
\end{equation}
\noindent Using a single network allows the two tasks to share the first layers. This results in a more compact architecture, which is easier to deploy on embedded devices without sacrificing the network performance~\cite{10.1007/978-3-030-33676-9_3}.

The final network architecture (Fig.~\ref{fig:archi}) is composed of a context aggregation module, a regression module in charge of predicting the disparity residuals, and an outlier detection module. Monte-Carlo dropout layers are positioned in the intermediate layers of the network, while the variational inference layers are positioned at the end of the network.

\subsubsection{Context aggregation module}
\label{sec:ourapproach:functional:ctx}

\begin{figure}[tbh]
    \centering
    \includegraphics[width=0.48\textwidth]{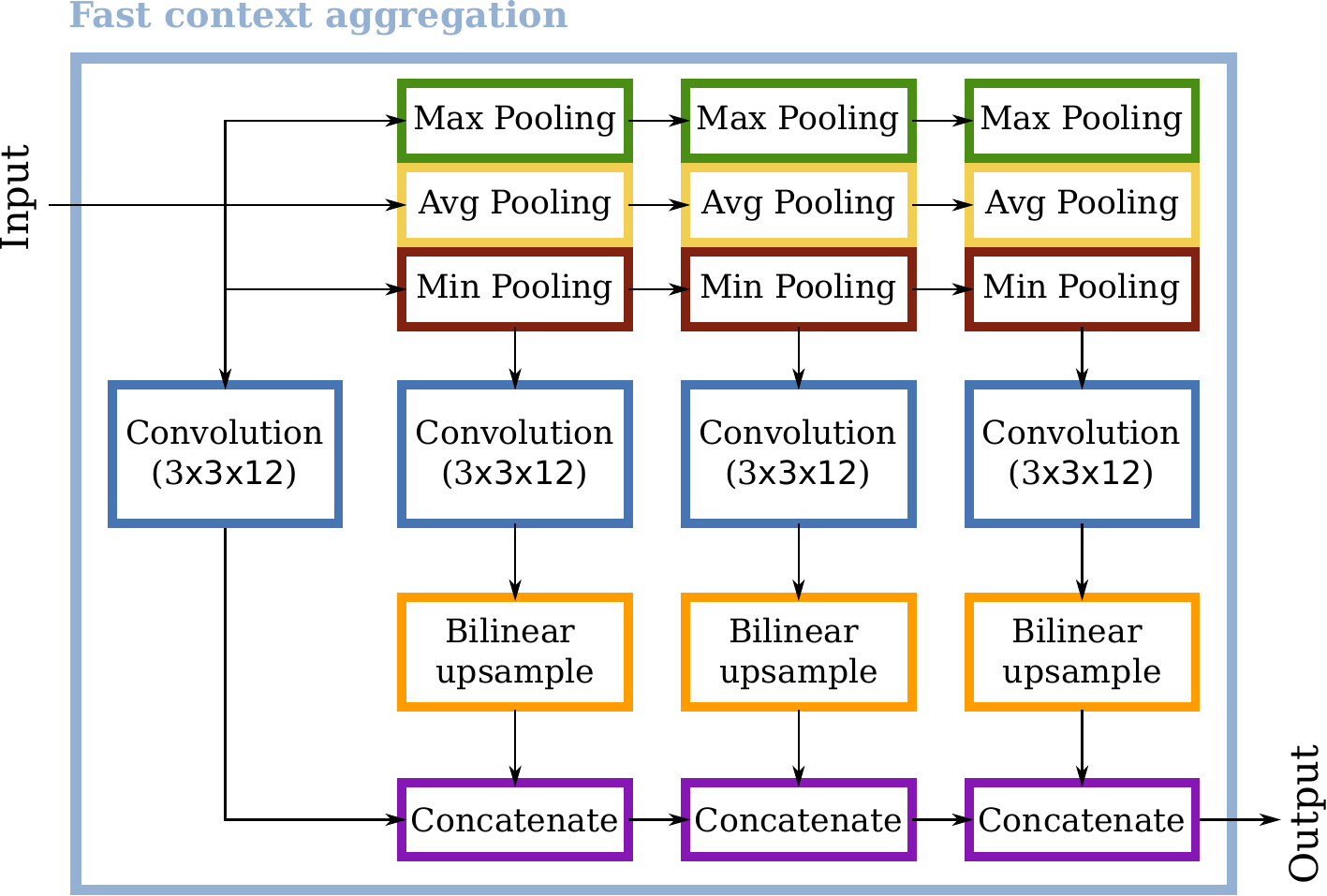}
    \caption{The fast context aggregation layer inner architecture}
    
    \label{fig:fast_context_aggregation}
\end{figure}

Aggregating non-local information is key to the performance of fully-convolutional networks. An hourglass model is typically used for this where the inner layers are pooled to reduce the resolution while the number of channels is increased to retain the information. This is prohibitively expensive for ultra-light networks. In mitigation, the inner levels are bypassed and each tensor is transferred to the levels on the other side of the hourglass. Thus, the required number of channels in the inner layers can be drastically reduced while maintaining high-resolution information. We extend this concept and propose a fast context aggregation layer to extract contextual information with a minimum number of convolutions (Fig.~\ref{fig:fast_context_aggregation}). Three types of pooling algorithms are used at each level in this architecture: max pooling, average pooling, and min pooling. These enable the network to estimate a rough distribution of the features within each pooled window. A single convolution layer is then applied at each level before the images are directly upsampled and concatenated together. Although this approach may seem restrictive, it produces good results in the case of active stereo. This is because, unlike passive stereo, the surrounding of a match contains a lot of information, and hence it reduces the amount of global information that the model requires.

A single fast context aggregation layer is used in the context aggregation module. Its role is to aggregate context for both the subsequent disparity regression and outlier detection modules.

\subsubsection{Disparity regression module}

The disparity regression module is made up of a series of $1 \times 1$ convolution layers. The first two layers use the Leaky-ReLU activation function with a negative slope of $0.3$. This function preserves most of the benefits of ReLU activation while limiting the risk of certain neurons dying out \cite{2020DyingReLU}. This is useful for light architectures which do not have the level of redundancy present in larger networks. After those two layers, a branch shares the features used for regression with the outlier detection module (Fig.~\ref{fig:archi}). This is meant to promote more co-evolution with the outlier detection module.

The second to last layers use a more advanced activation function, which we refer to as Leaky-Hardswish. This is a variant of the Hardswish function introduced in MobileNetv3 \cite{howard2019searching}, which is defined as:
\begin{equation}
   f(x) = x \cdot \max\left(0,\min\left(1,\dfrac{x+1}{2}\right)\right) + \lambda x.
\end{equation}

\noindent The constant $\lambda$ is used to ensure that the derivative of $f$ is non-zero for the set of real numbers, minus singleton points. Although $f$ serves the same purpose as the Leaky-ReLu activation but is based on the Hardswish function, it has a quadratic region that can be exploited to smooth the output disparity. The final layer is a $1 \times 1$ convolution layer without an activation function. Its purpose is to aggregate all the channels from the previous layer into the final estimate for $\estimatordisprefiner$. Following Equation~\eqref{eq:dispfromraw}, the final disparity estimate is obtained by taking the sum of $\rawdisparity$ and $\estimatordisprefiner$.

\subsubsection{Outliers detection module.} The outlier detection module is based on the same principles as the disparity regression module, with a series of $1 \times 1$ convolution layers acting on the previously aggregated context. The Leaky-ReLu activation function is used again, but with a smaller slope of $0.1$ for negative activations instead of $0.3$. The rationale behind this decision is to provide stronger non-linearities for the module used to perform classification instead of regression.

Our experiments have also shown that the outlier classification module tends to be over-pessimistic and might, during training, start to classify all pixels as outliers if only a few cannot be differentiated from inliers. Those problematic pixels are usually located close to the boundaries between inliers and outliers. To reduce the chances of this happening, the final convolution layer in the module has a receptive field of $5 \times 5$ instead of $1 \times 1$. To keep the number of parameters low, the filter has been implemented as a separable convolution \cite{SeparableConvolutionsStereo}, with the first layer aggregating the $25$ inputs channels, then one layer applying a $1 \times 5$ filter on the resulting channel, followed by a similar layer of shape $5 \times 1$. It means $35$ parameters are required instead of $625$. The final output is then passed through a sigmoid activation function to get $p_{out}$.

Our final architecture is very minimalistic, making it more suitable for embedded systems with limited memory and computational power than the state-of-the-art models (see Table~\ref{table:modelfoootprint}).


\section{Experiments}
\label{sec:evaluation}

\begin{figure}[bt]
    \centering
    \begin{subfigure}[t]{0.154\textwidth}
    \includegraphics[height=1.7cm]{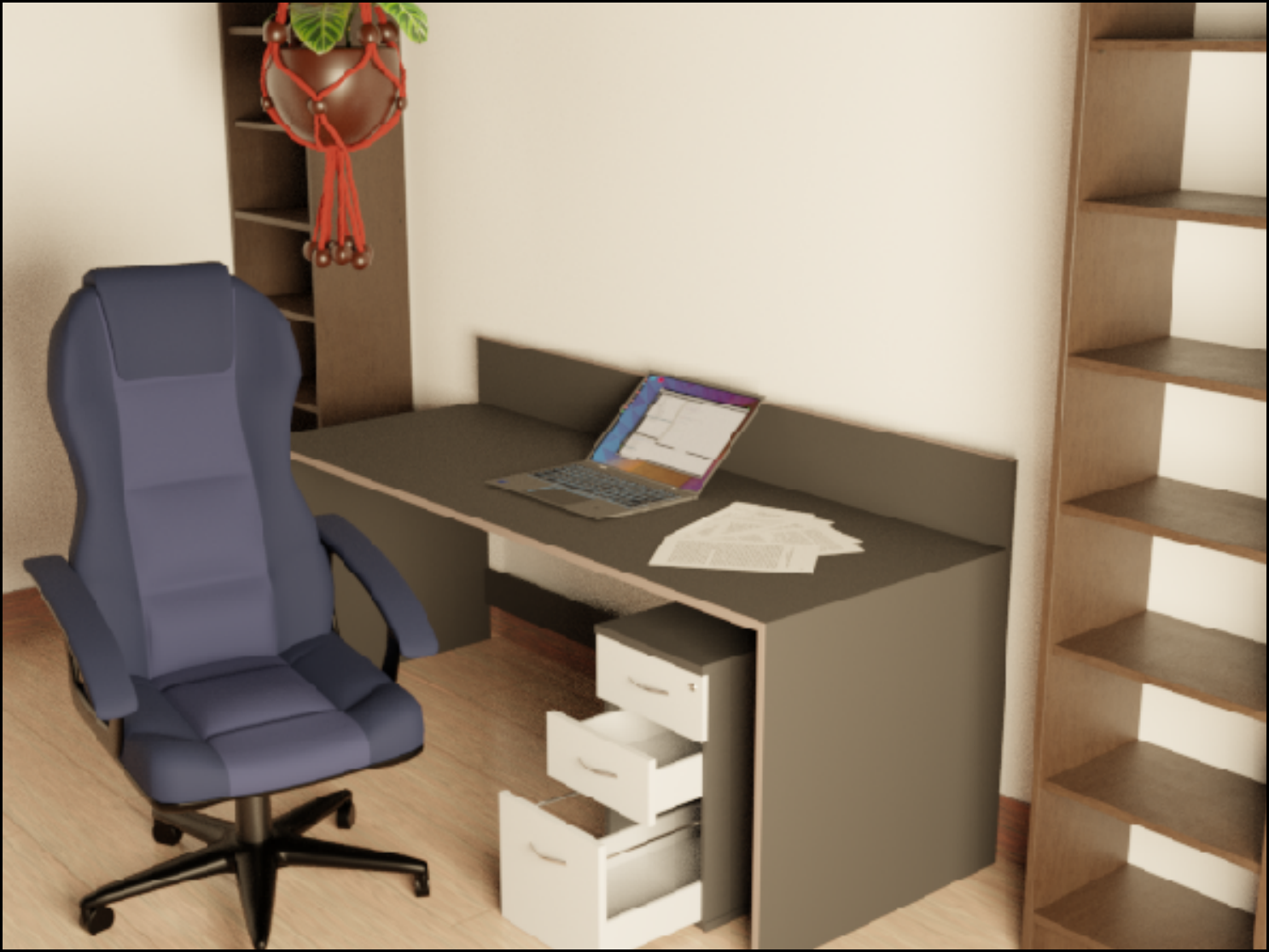}
    \caption{}
    \label{fig:model_inputs:rgb}
    \end{subfigure}
    \begin{subfigure}[t]{0.154\textwidth}
    \includegraphics[height=1.7cm]{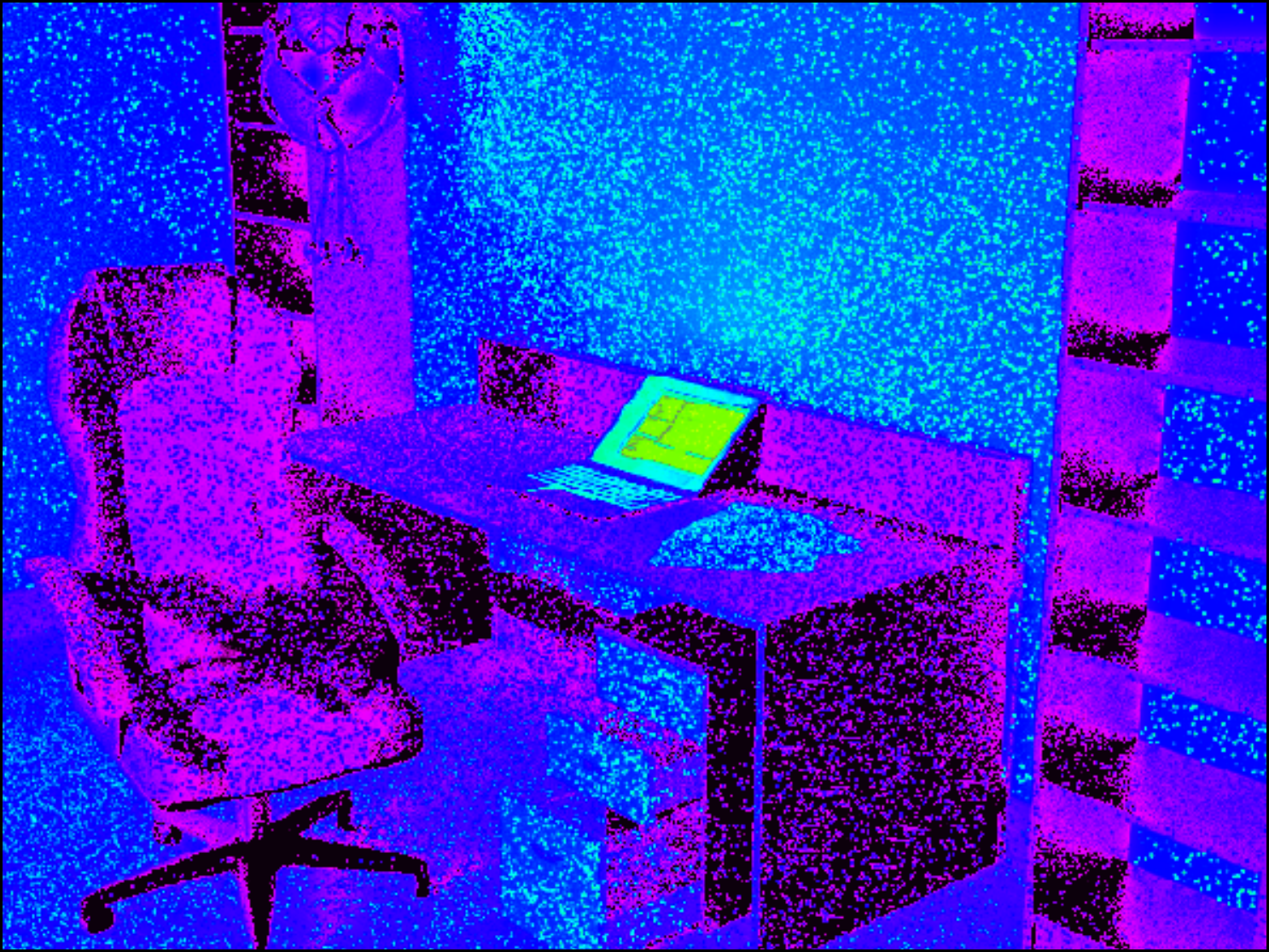}
    \caption{}
    \label{fig:model_inputs:snir}
    \end{subfigure}
    \begin{subfigure}[t]{0.154\textwidth}
    \includegraphics[height=1.7cm]{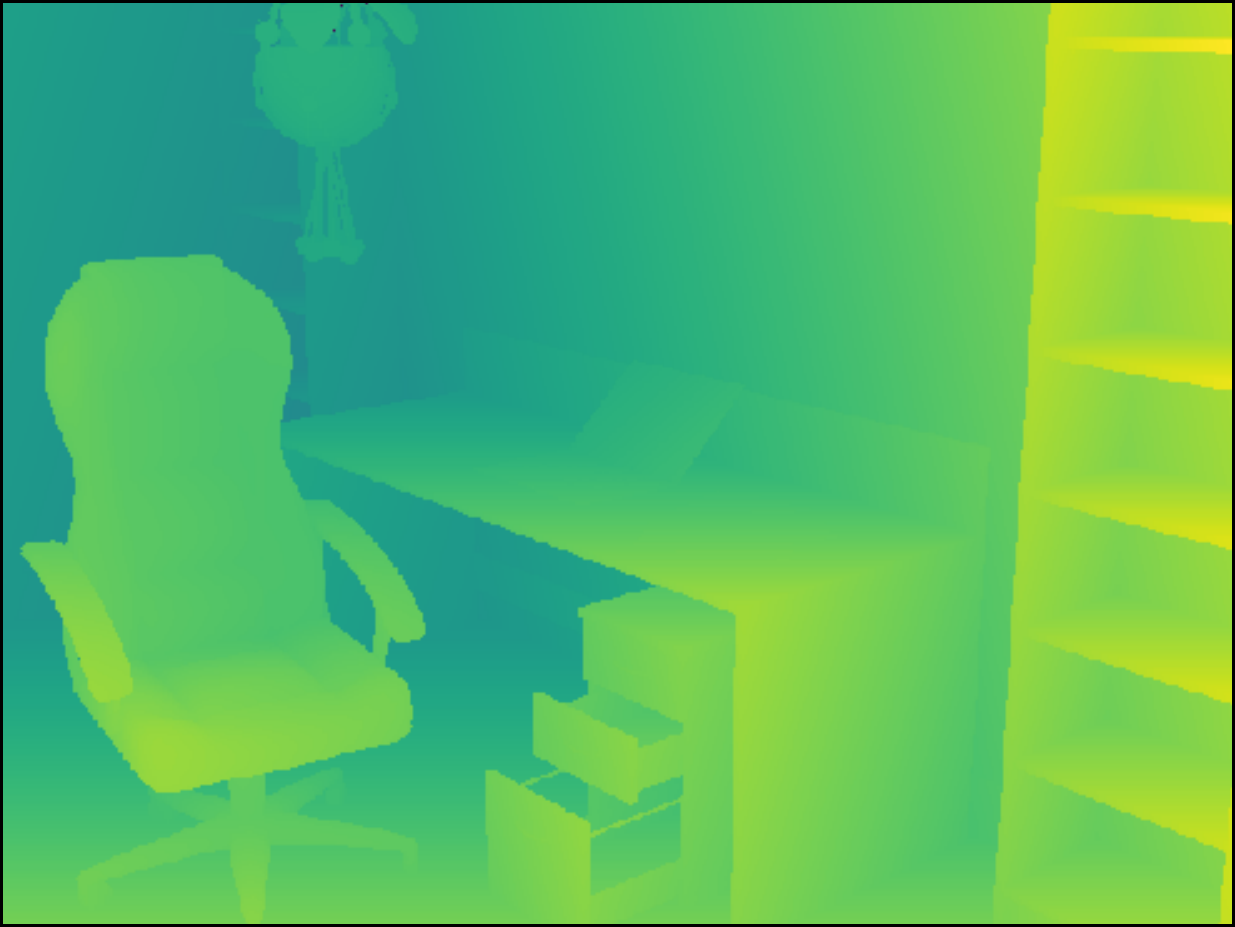}
    \caption{}
    \label{fig:model_inputs:tdisp}
    \end{subfigure}
    
    \begin{subfigure}[t]{0.158\textwidth}
    \includegraphics[height=1.7cm]{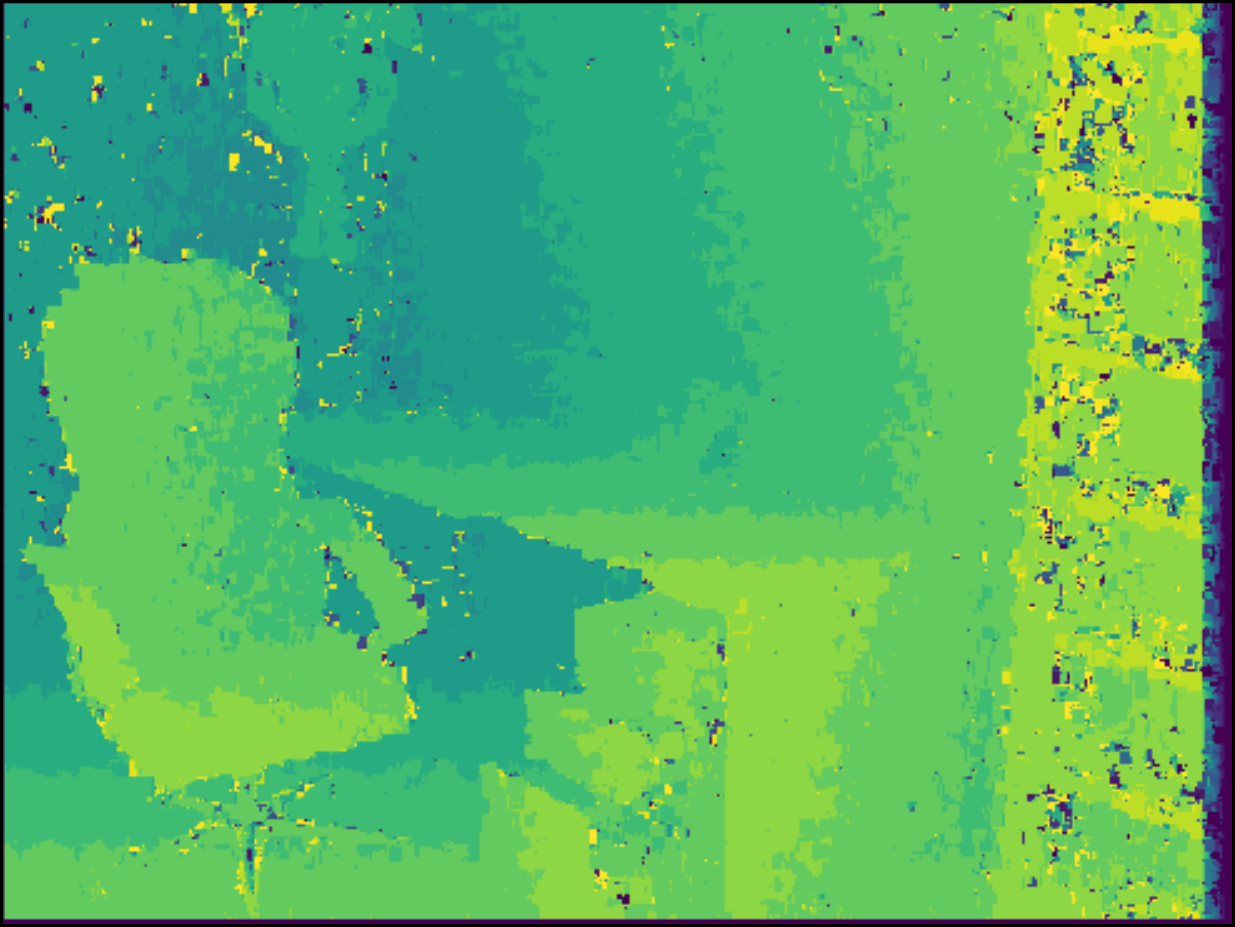}
    \caption{}
    \label{fig:model_inputs:disp}
    \end{subfigure}
    \begin{subfigure}[t]{0.158\textwidth}
    \includegraphics[height=1.7cm]{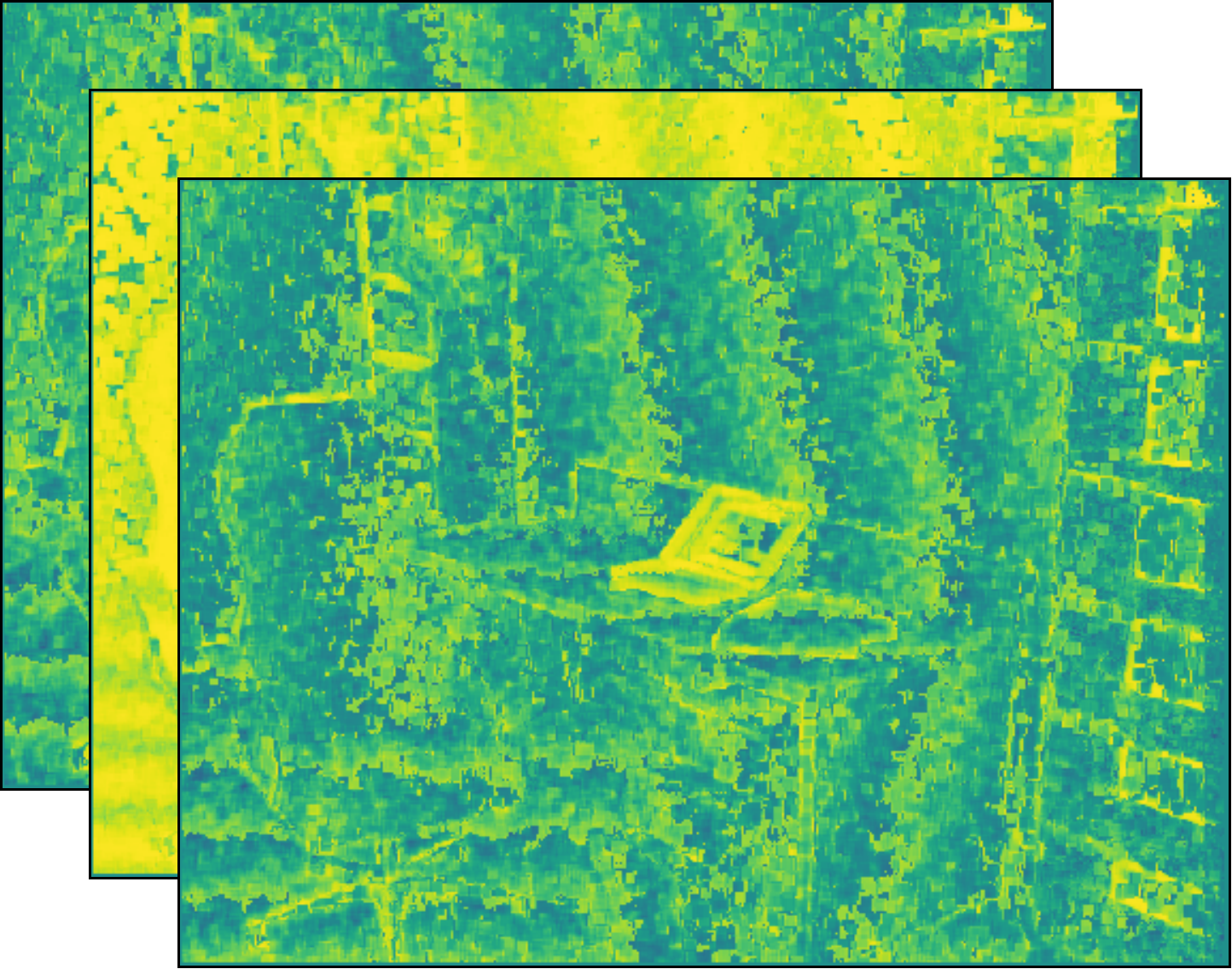}
    \includegraphics[height=1.7cm]{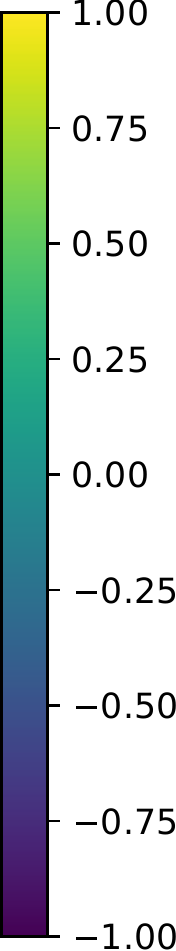}
    \caption{}
    \label{fig:model_inputs:cv}
    \end{subfigure}
    \begin{subfigure}[t]{0.15\textwidth}
    \includegraphics[height=1.7cm]{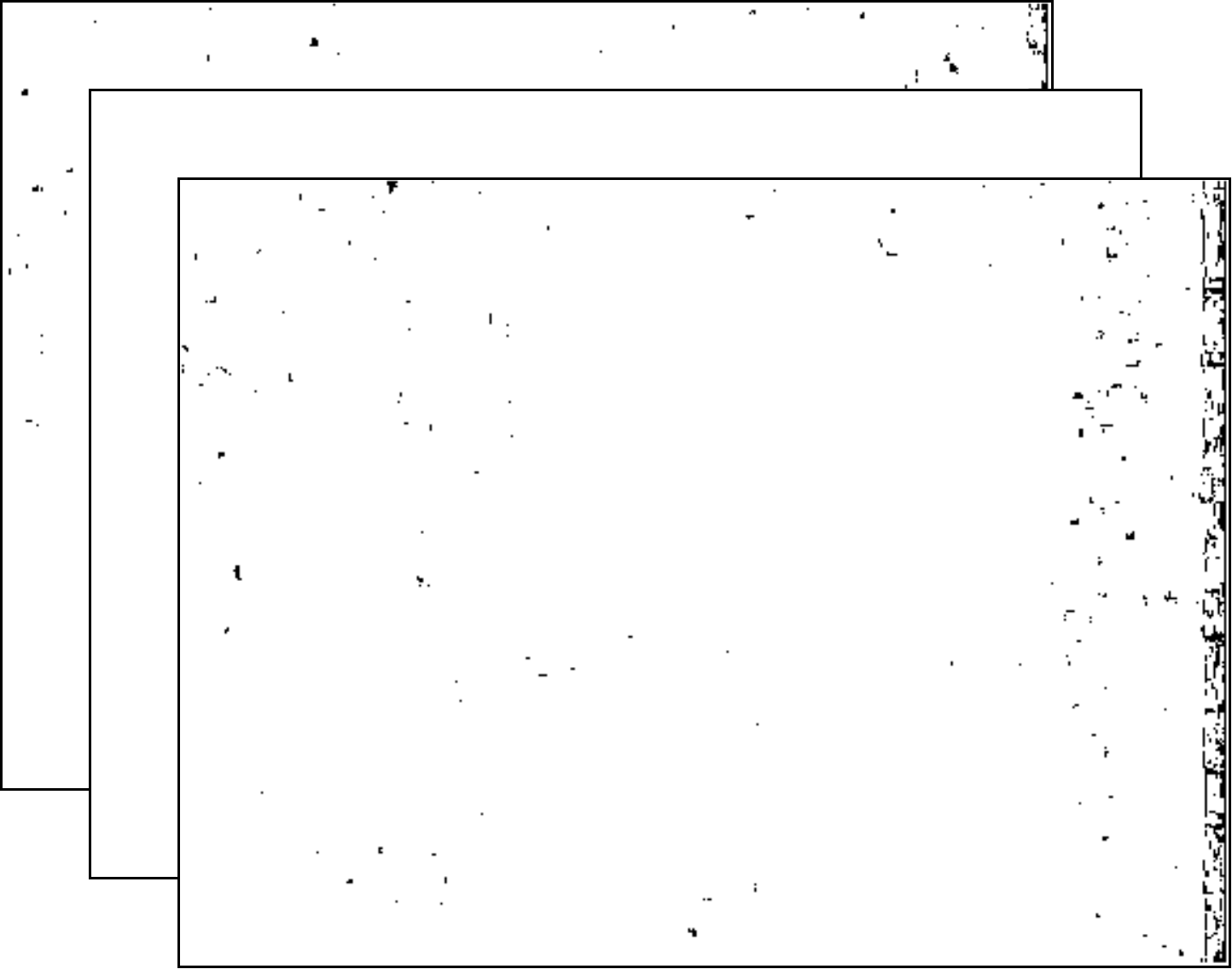}
    \caption{}
    \label{fig:model_inputs:ib}
    \end{subfigure}
    
    
    \caption{The \ourdatasetname{} dataset is composed of different scenes with (\subref{fig:model_inputs:rgb}) RGB frames, (\subref{fig:model_inputs:snir}) NIR frames with pseudo-random pattern and  (\subref{fig:model_inputs:tdisp}) ground truth disparity used to supervise the regression module. An initial algorithm, NCC, extracts the raw disparity (\subref{fig:model_inputs:disp})which will serve as inputs of the BNN , (\subref{fig:model_inputs:cv}) the truncated cost volume and (\subref{fig:model_inputs:ib}) the support indicator function.}
    \label{fig:model_inputs}
\end{figure}

\begin{figure}[tb]
    \centering
    \begin{subfigure}[b]{0.22\textwidth}
    \includegraphics[width=\textwidth]{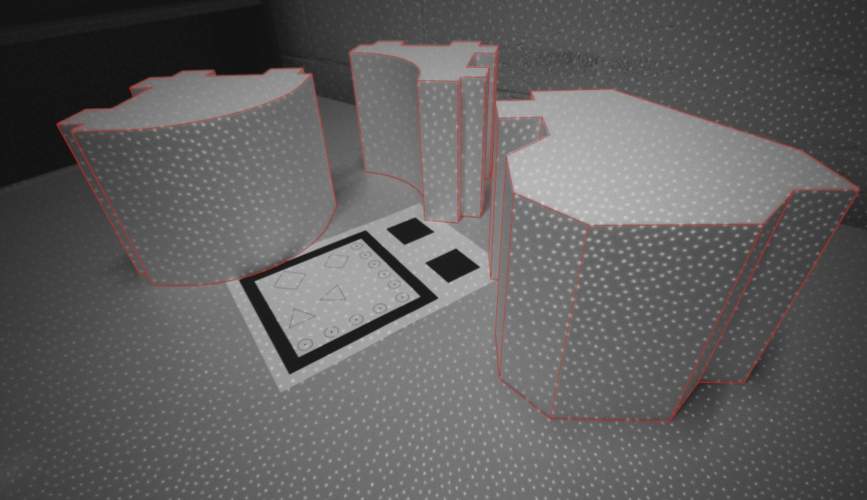}
    \caption{}
    \label{fig:realsenseimgpair:left}
    \end{subfigure}
    \hfill
    \begin{subfigure}[b]{0.22\textwidth}
    \includegraphics[width=\textwidth]{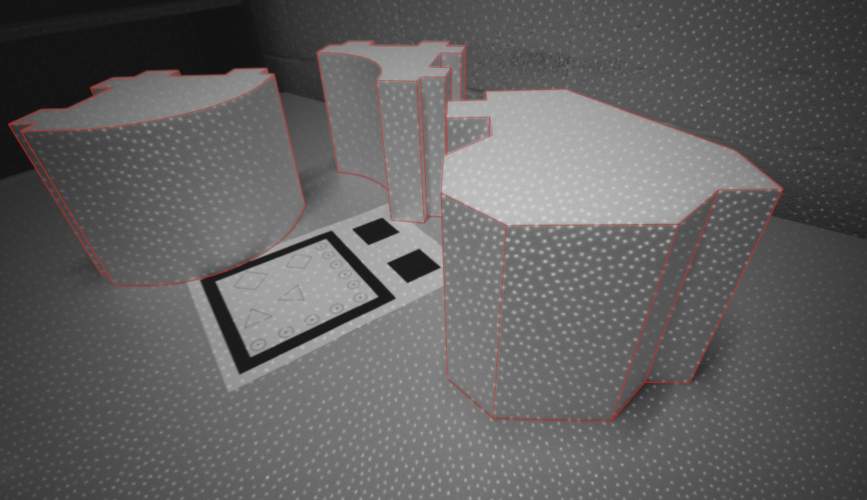}
    \caption{}
    \label{fig:realsenseimgpair:right}
    \end{subfigure}
    
    \caption{Left (\subref{fig:realsenseimgpair:left}) and right  (\subref{fig:realsenseimgpair:right}) frames of an Image pair from the Shapes dataset with the outlined CAD models aligned onto the real scene.}
    \label{fig:realsenseimgpair}
\end{figure}

\begin{figure*}[t]
    \centering
    
    \begin{subfigure}[b]{0.45\textwidth}
    \includegraphics[width=\textwidth]{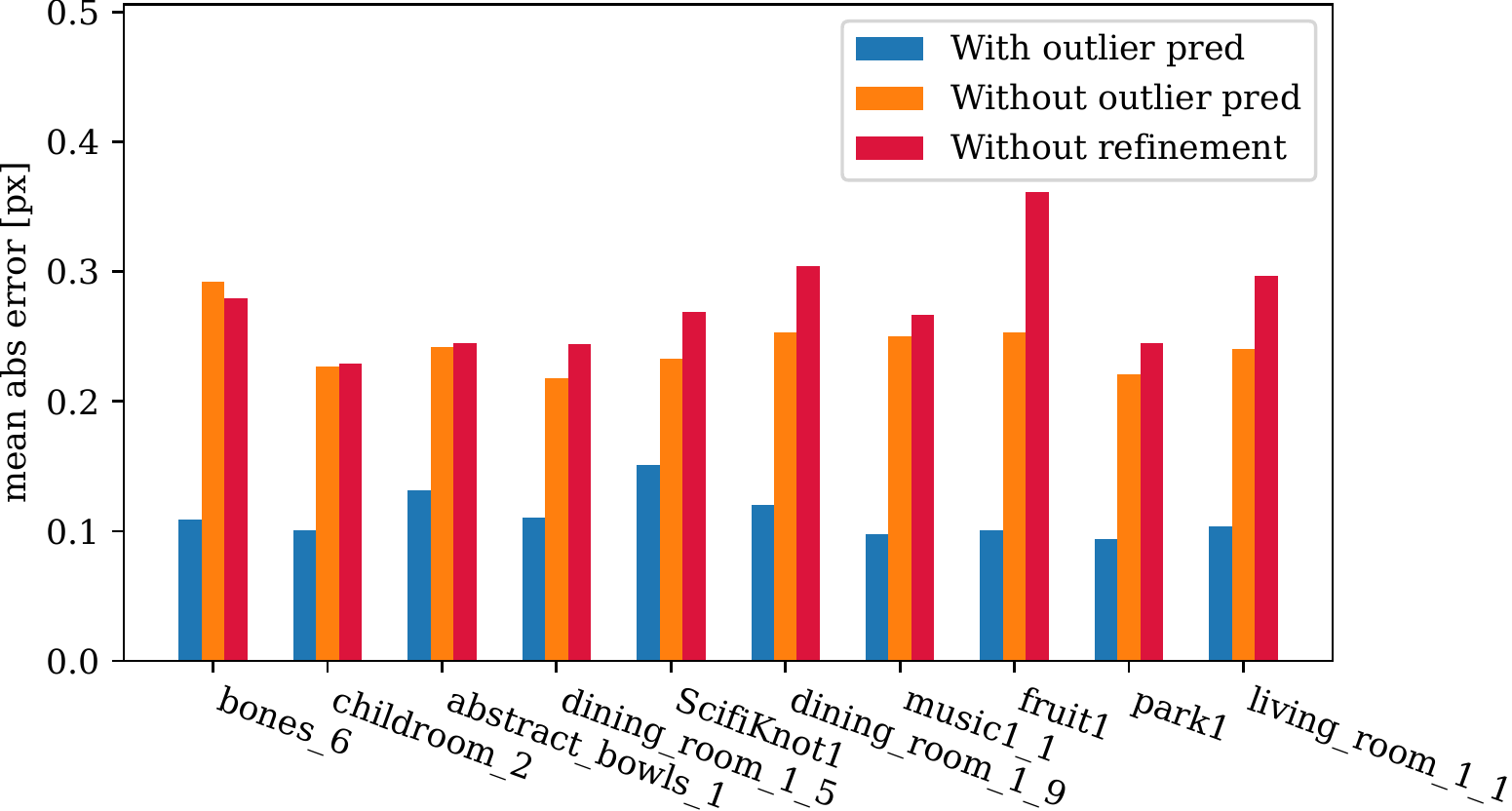}
    \caption{MAE computed on areas where $p_{out} \le 5\%$.}
    \label{fig:results:meanabserror}
    \end{subfigure}
    \hfill
    \begin{subfigure}[b]{0.45\textwidth}
    \includegraphics[width=\textwidth]{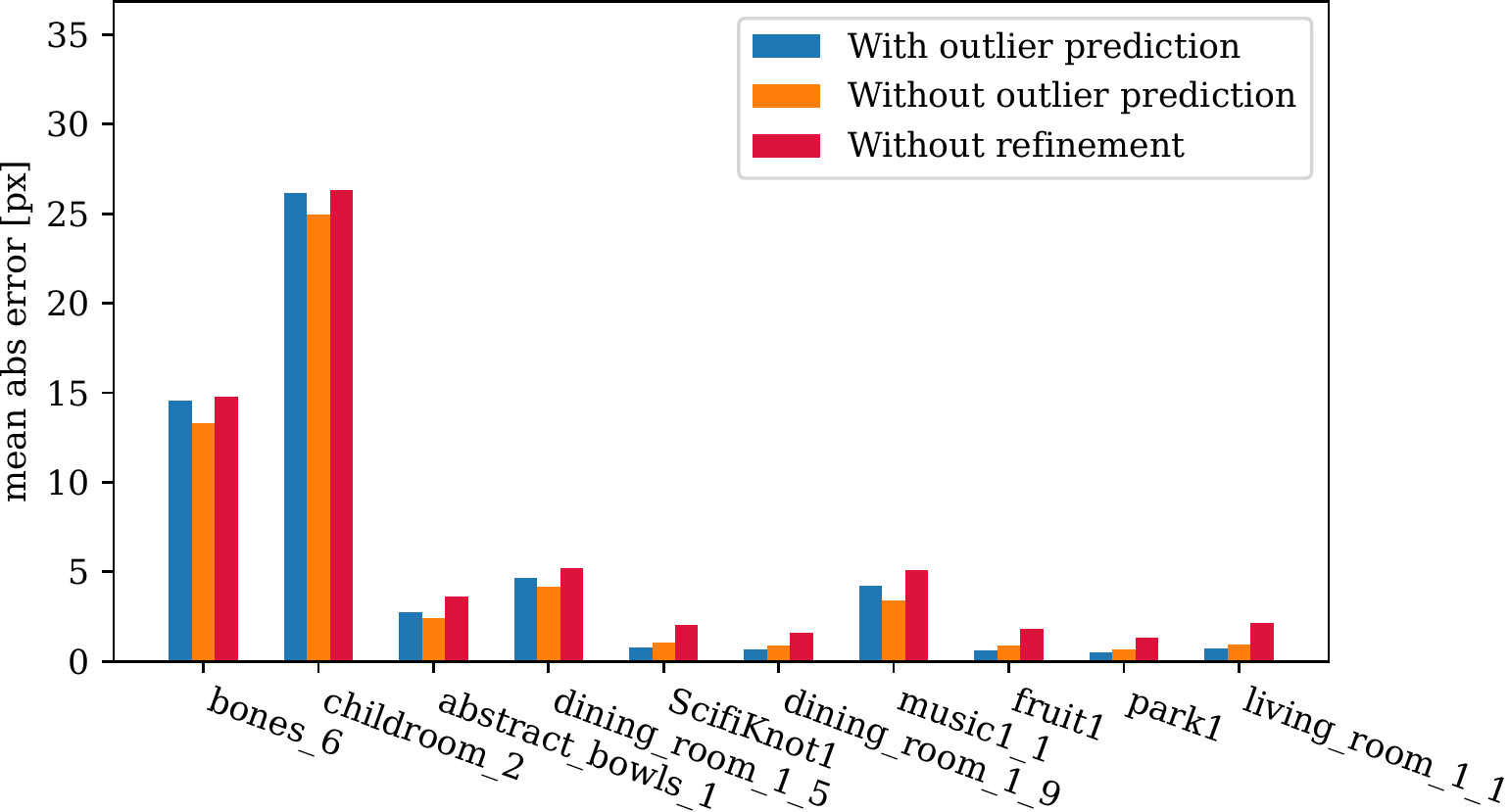}
    \vspace{-17pt}
    \caption{MAE computed over the entire image.}
    \label{fig:results:meanabserror_global}
    \end{subfigure}
    
    
    \caption{MAE of the proposed model on a selection of images from the \ourdatasetname{} \cite{ourDataset} test set, with and without outlier prediction, compared against the raw disparity for the selected inliers.}
    
    \label{fig:results}
\end{figure*}

We trained the proposed model using the active stereo frames from the \ourdatasetname{} dataset \cite{ourDataset}, a simulated dataset containing raw active stereo frames annotated with high-quality ground truth suitable for subpixel accurate models. We then evaluated our model on both the \ourdatasetname{} dataset and images from the Shapes dataset. The Shapes dataset is a dataset of real active stereo images we acquired with an Intel RealSense camera. It contains images of polystyrene 3D blocks cut from Computer-Aided Design (CAD) models with known dimensions. The CAD models can then be realigned on the images to obtain a highly accurate ground truth disparity map over the shapes. More details on this dataset are given in the Supplementary Material. In Section~\ref{sec:evaluation:ablation} we discuss the effect of our Bayesian training strategy by comparing our model trained with or without the outlier identification module. In Section~\ref{sec:evaluation:comp}, we compare our method with (1) a selection of state-of-the-art active and passive stereo models, (2) real-time optimized models, and (3) baseline methods.
 We also tested the performances of our model on the images from the Middlebury 2014 test set~\cite{10.1007/978-3-319-11752-2_3} to check if the model is able to generalize to passive stereo. The details of that experiment are discussed in the Supplementary Material.

\begin{figure}[tb]
    \centering
    \includegraphics[width=0.45\textwidth]{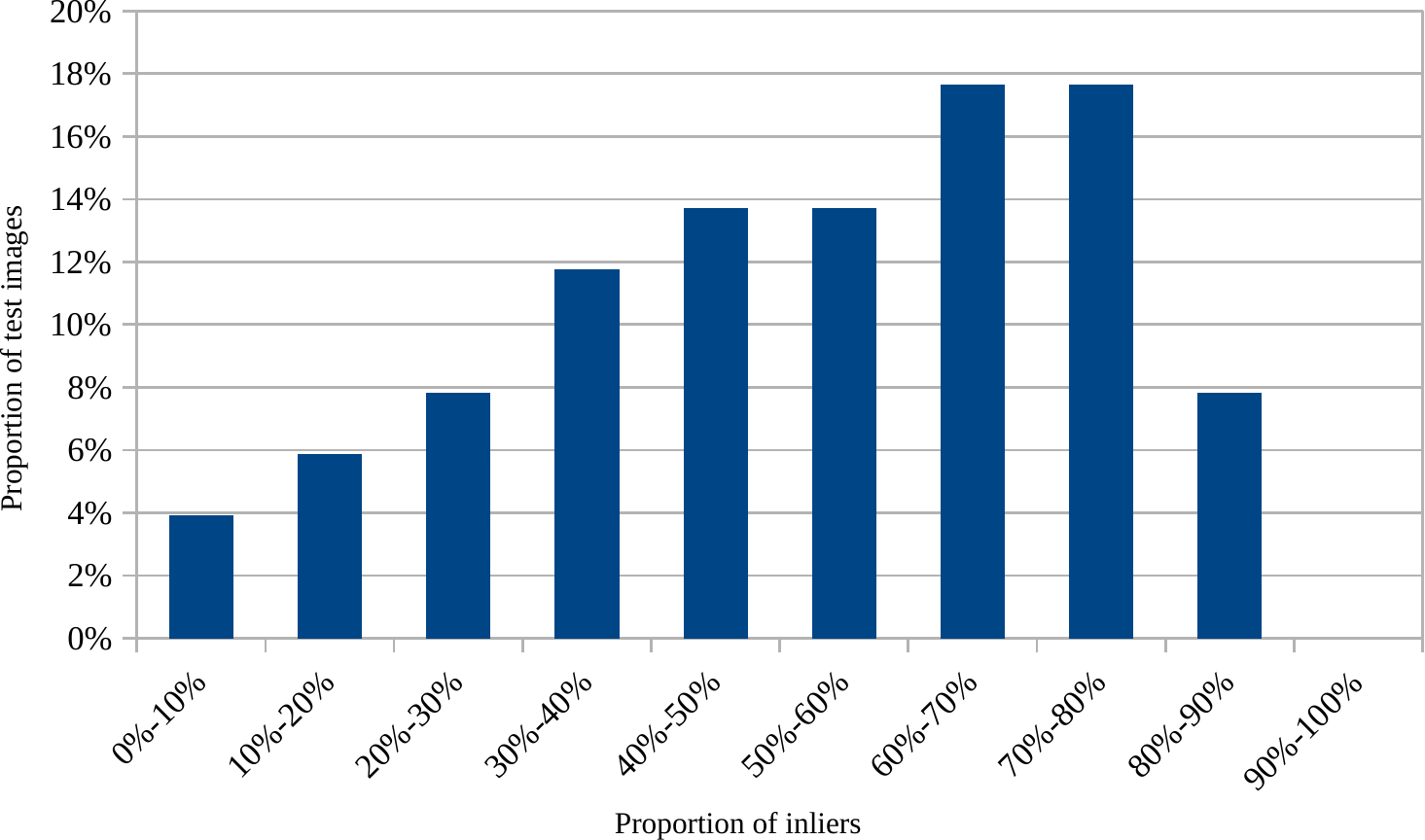}
    \caption{Distribution of the proportion of inlier pixels for the images in the \ourdatasetname{} test set \cite{ourDataset}.}
    \label{fig:prop_inliers}
\end{figure}

\subsection{Ablation study}
\label{sec:evaluation:ablation}

\begin{figure*}[ptb]
    
    \begin{subfigure}[t]{0.3\textwidth}
        \includegraphics[width=\textwidth]{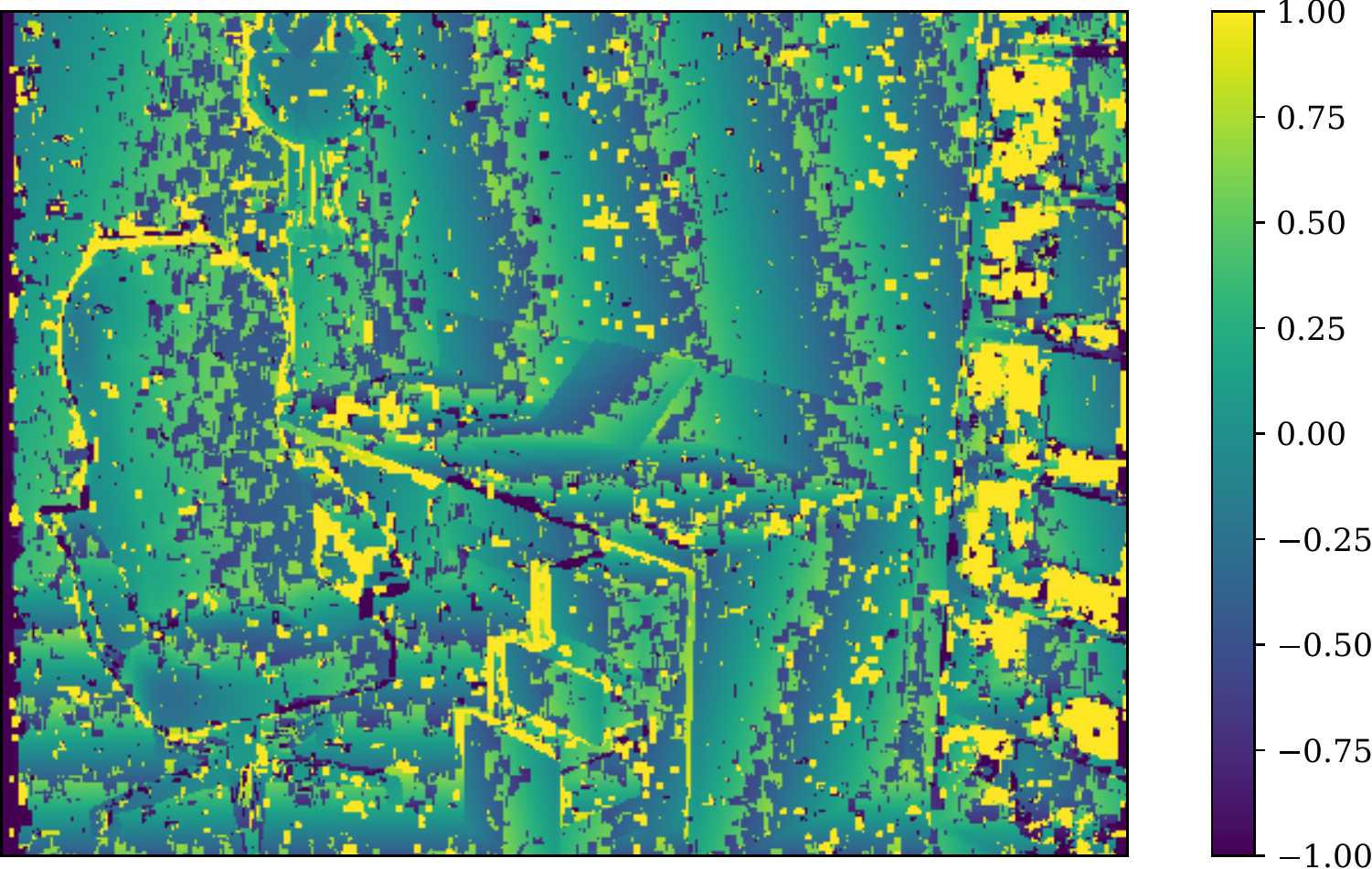}
        \caption{Error for raw disparity [px].}
        \label{fig:visualresults:rawdisperror}
    \end{subfigure}
    \hfill
    \begin{subfigure}[t]{0.3\textwidth}
        \includegraphics[width=\textwidth]{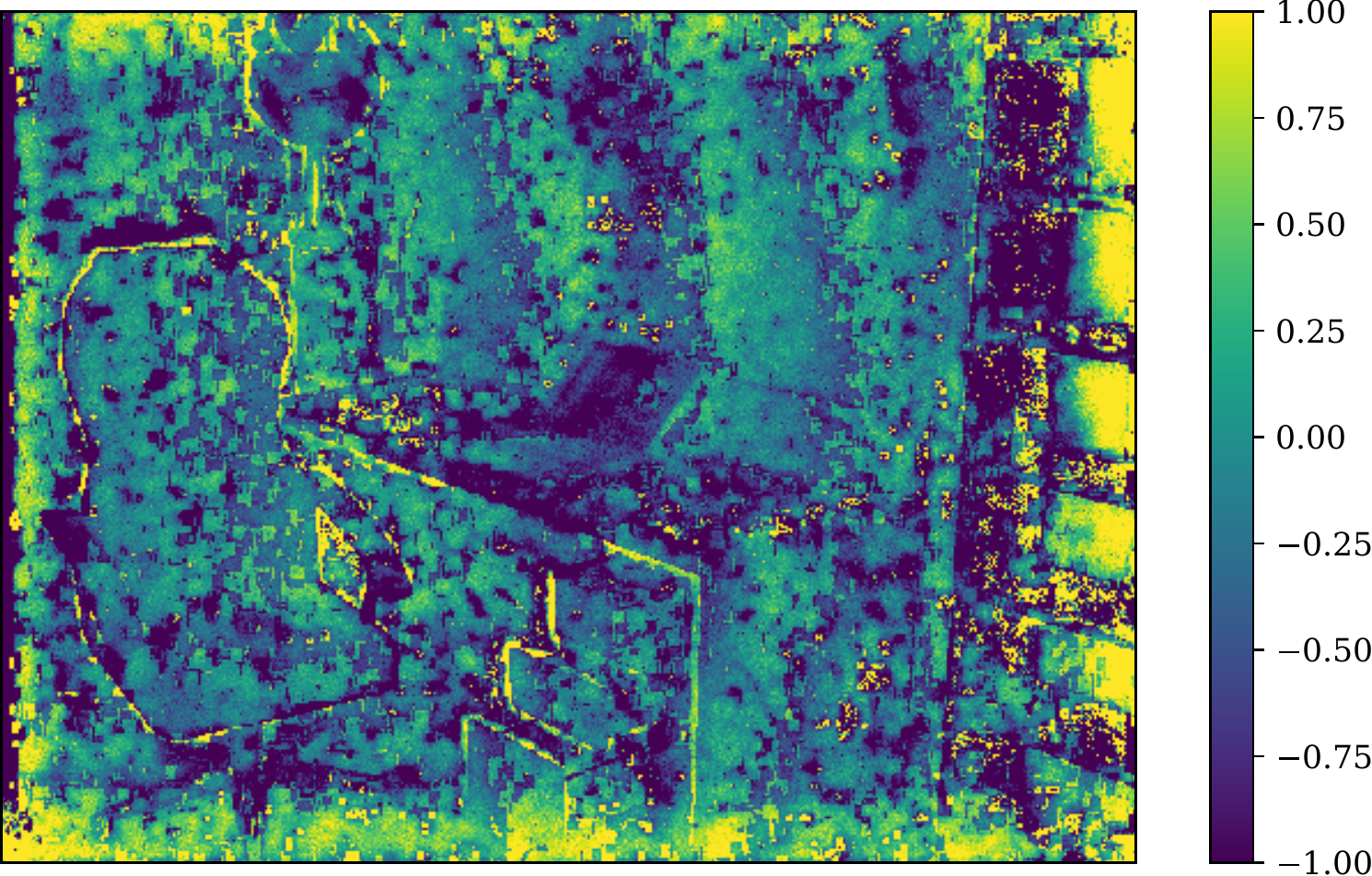}
        \caption{Error for refined disparity [px], without outliers module.}
        \label{fig:visualresults:refineddispwithoutoutlierserror}
    \end{subfigure}
    \hfill
    \begin{subfigure}[t]{0.3\textwidth}
        \includegraphics[width=\textwidth]{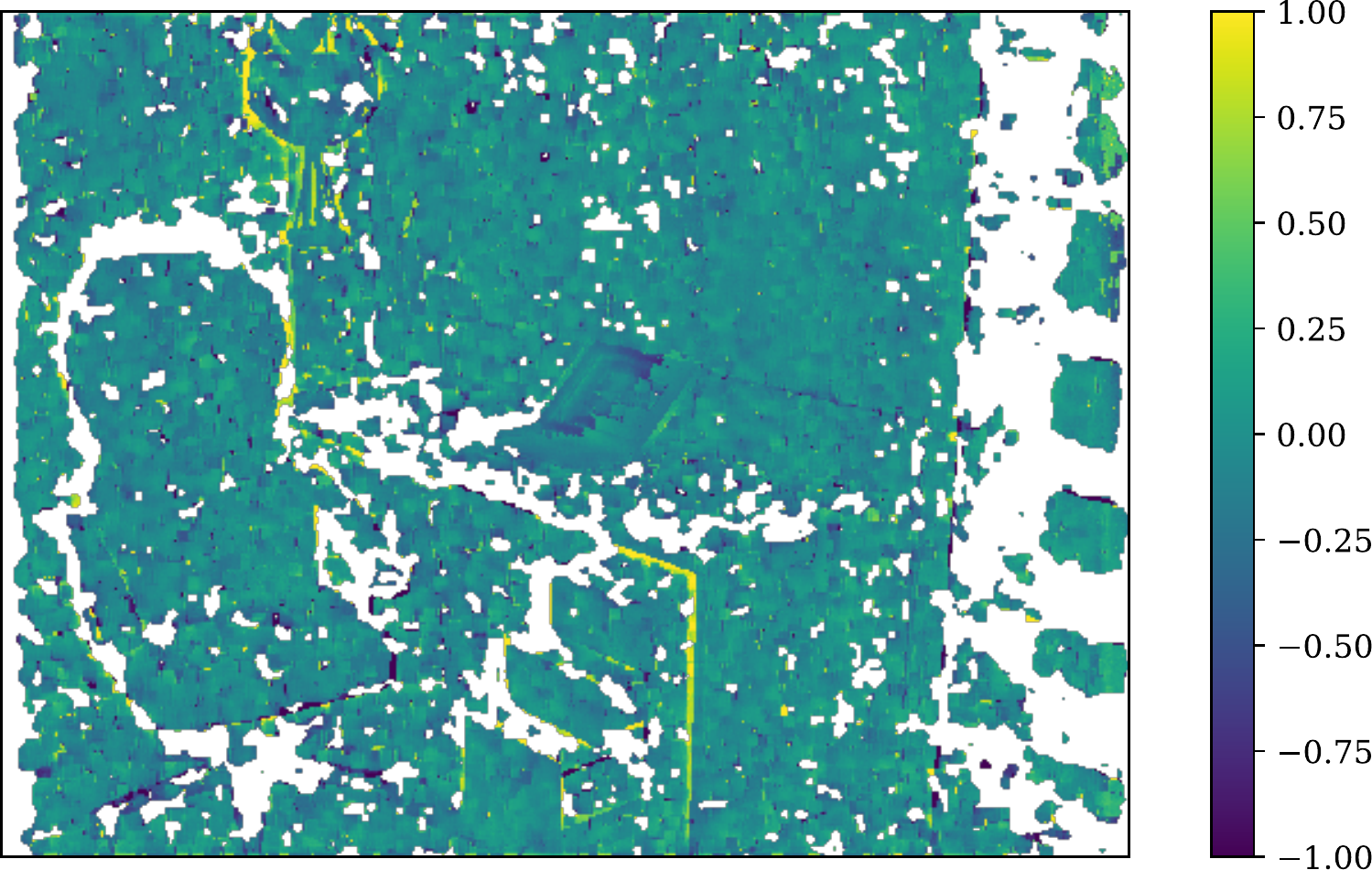}
        \caption{Error for refined disparity [px], with outliers removal module.}
        \label{fig:visualresults:refineddispwithoutlierserror}
    \end{subfigure}
    
    \vspace{5pt}
    
    \begin{subfigure}[t]{0.3\textwidth}
        \includegraphics[width=\textwidth]{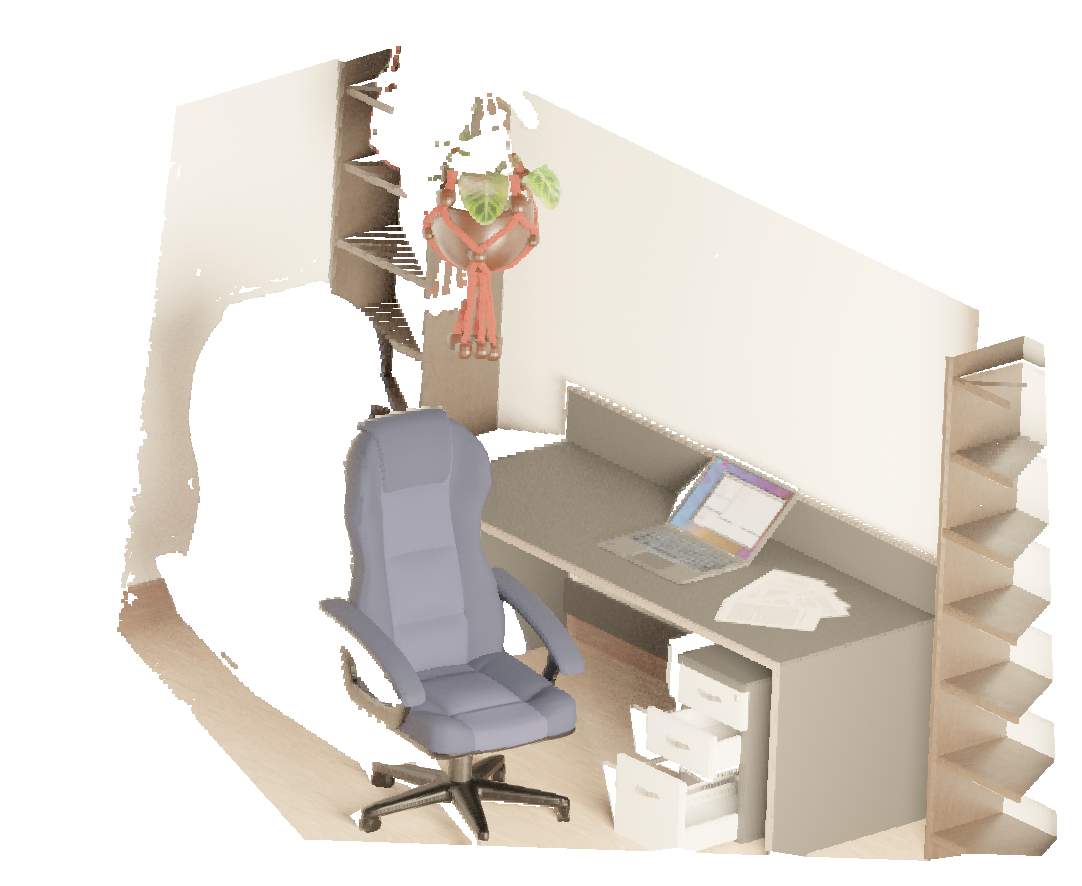}
        \caption{Ground truth point cloud.}
        \label{fig:visualresults:gt_point_cloud}
    \end{subfigure}
    \hfill
    \begin{subfigure}[t]{0.3\textwidth}
        \includegraphics[width=\textwidth]{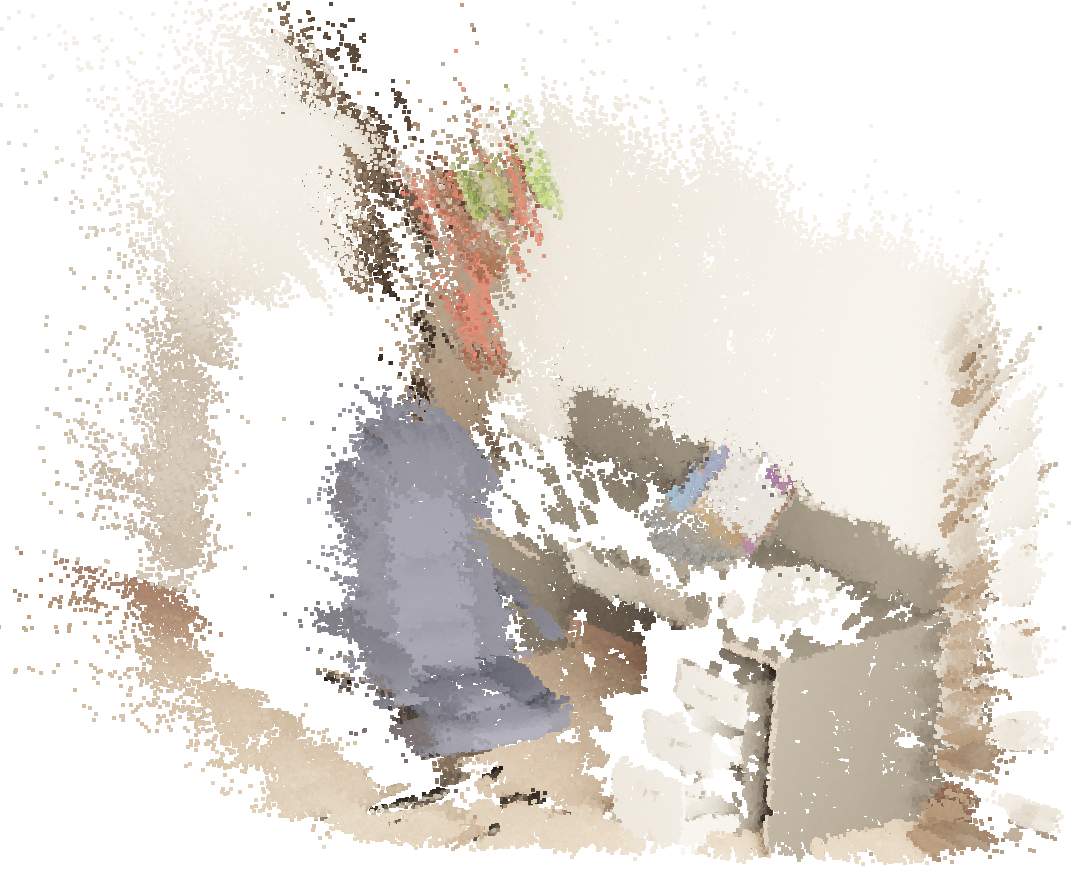}
        \caption{Point cloud of validated pixels without the outlier removal module.}
        \label{fig:visualresults:refined_without_outlier_point_cloud}
    \end{subfigure}
    \hfill
    \begin{subfigure}[t]{0.3\textwidth}
        \includegraphics[width=\textwidth]{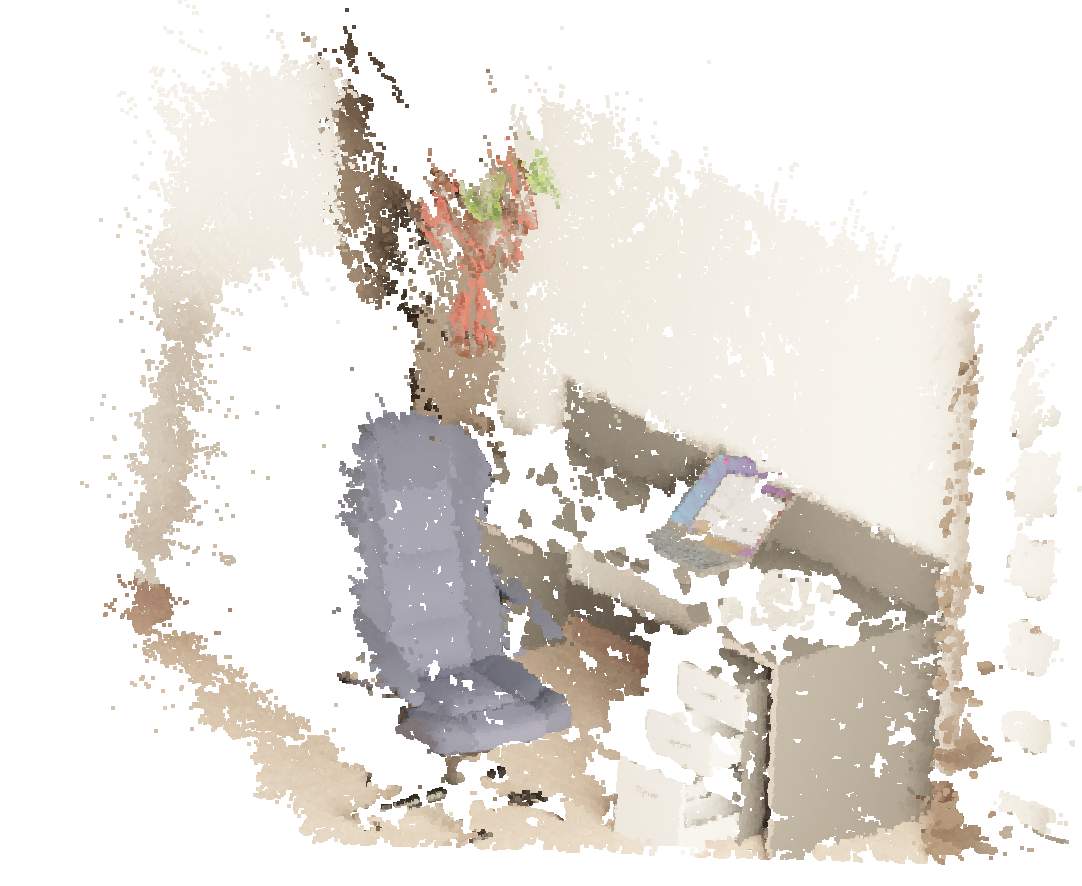}
        \caption{Point cloud of validated pixels with the outlier removal module.}
        \label{fig:visualresults:refined_with_outlier_point_cloud}
    \end{subfigure}
    
    \vspace{5pt}
    
    \caption{Our method allows to obtain visible improvements in disparity, and thus depth, reconstruction when training the neural network by focusing in priority on inliers.}
    \label{fig:visualresults}
    
\end{figure*}

\begin{figure*}[pt]
    \centering
    \begin{subfigure}[t]{0.3\textwidth}
    \includegraphics[width=\textwidth]{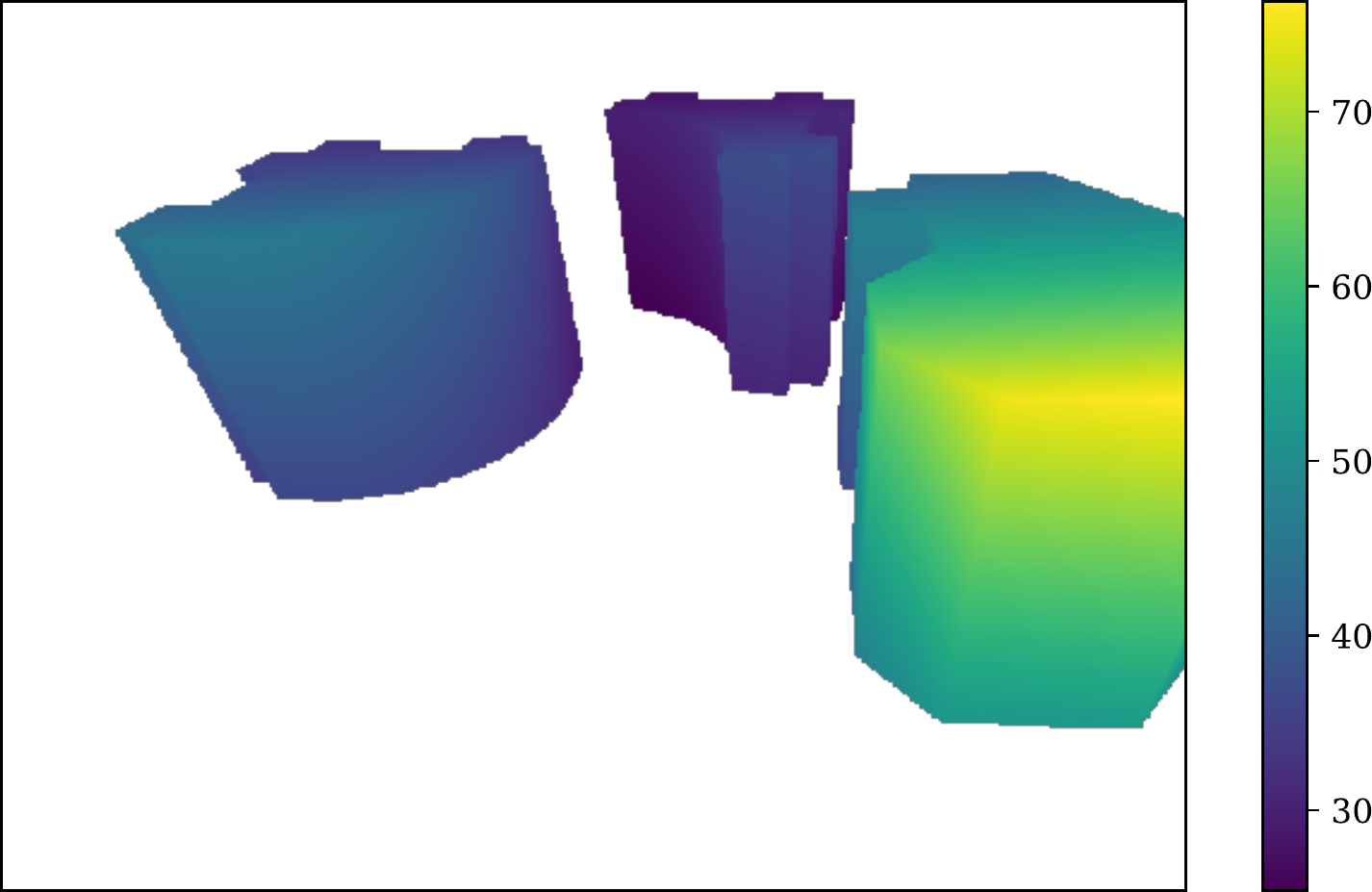}
    \caption{Ground truth.}
    \label{fig:resultsshapes:gt}
    \end{subfigure}
    \hfill
    \begin{subfigure}[t]{0.3\textwidth}
    \includegraphics[width=\textwidth]{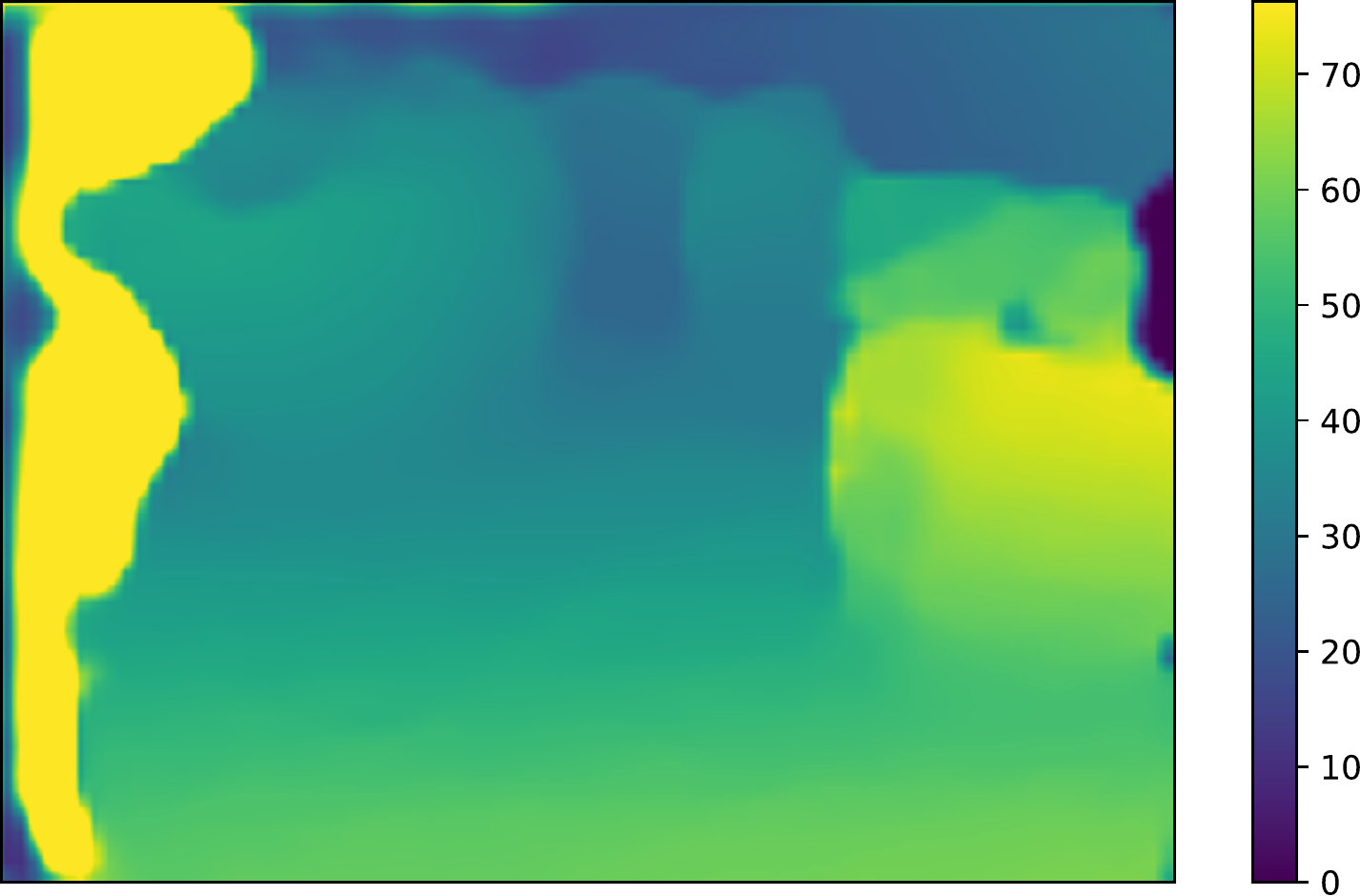}
    \caption{ActiveStereoNet self-supervised.}
    \label{fig:resultsshapes:asnselfsup}
    \end{subfigure}
    \hfill
    \begin{subfigure}[t]{0.3\textwidth}
    \includegraphics[width=\textwidth]{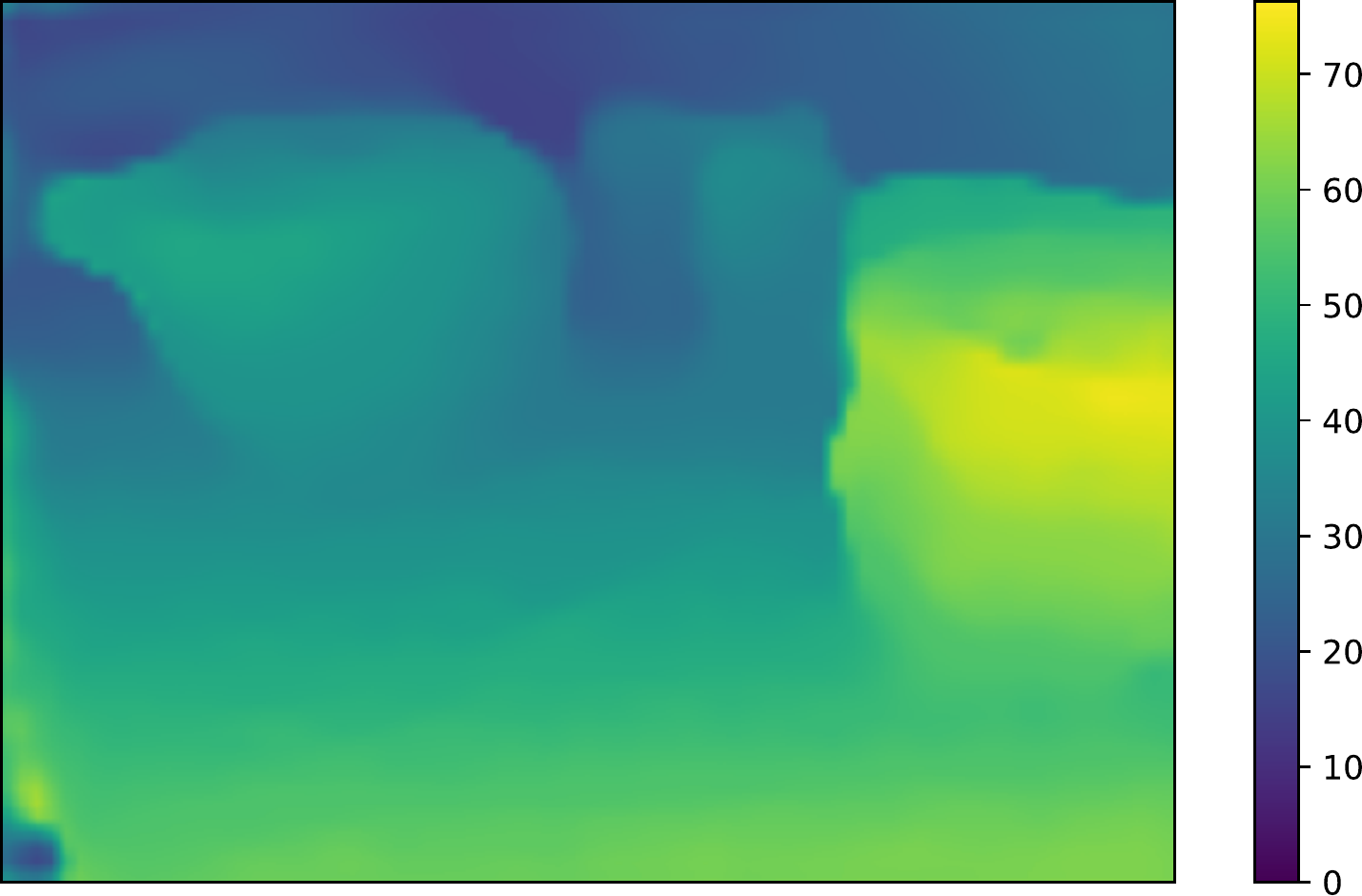}
    \caption{ActiveStereoNet finetuned.}
    \label{fig:resultsshapes:asnfinetuned}
    \end{subfigure}
    
    \vspace{5pt}
    
    \begin{subfigure}[t]{0.3\textwidth}
    \includegraphics[width=\textwidth]{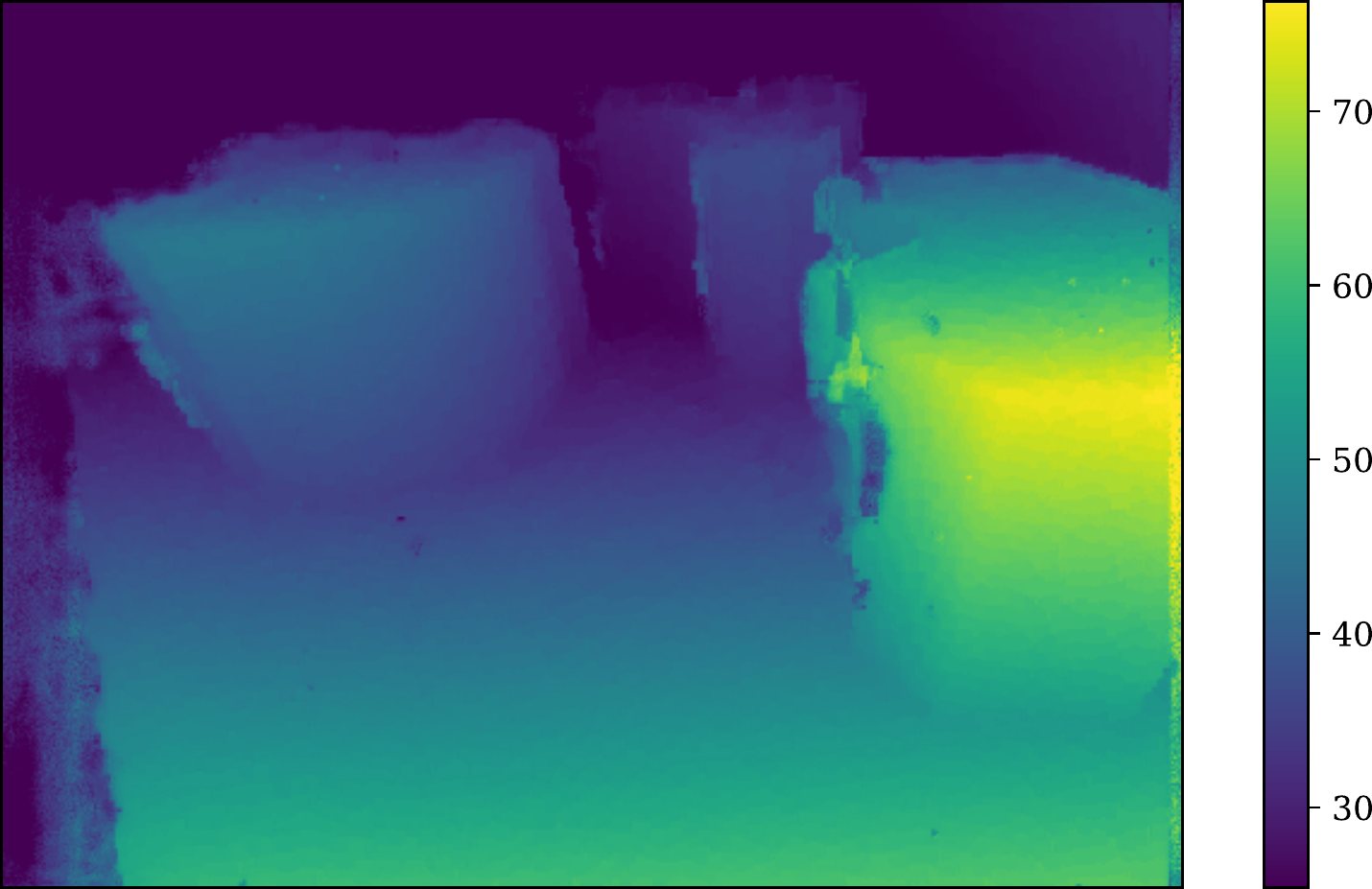}
    \caption{Ours (with the outliers module).}
    \label{fig:resultsshapes:ourwithoutliers}
    \end{subfigure}
    \hfill
    \begin{subfigure}[t]{0.3\textwidth}
    \includegraphics[width=\textwidth]{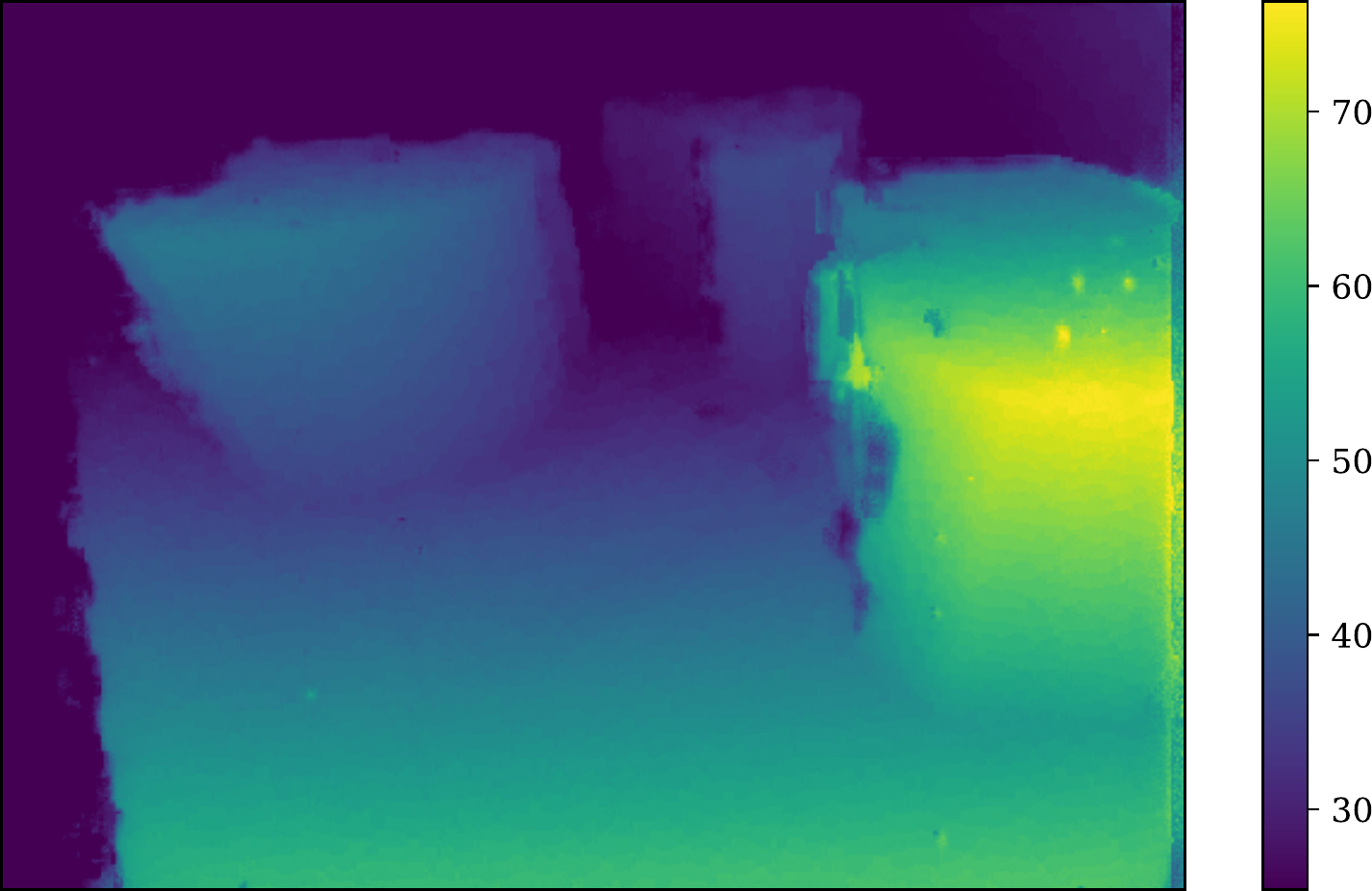}
    \caption{Ours (without the outliers module).}
    \label{fig:resultsshapes:ourwithoutoutliers}
    \end{subfigure}
    \hfill
    \begin{subfigure}[t]{0.3\textwidth}
    \includegraphics[width=\textwidth]{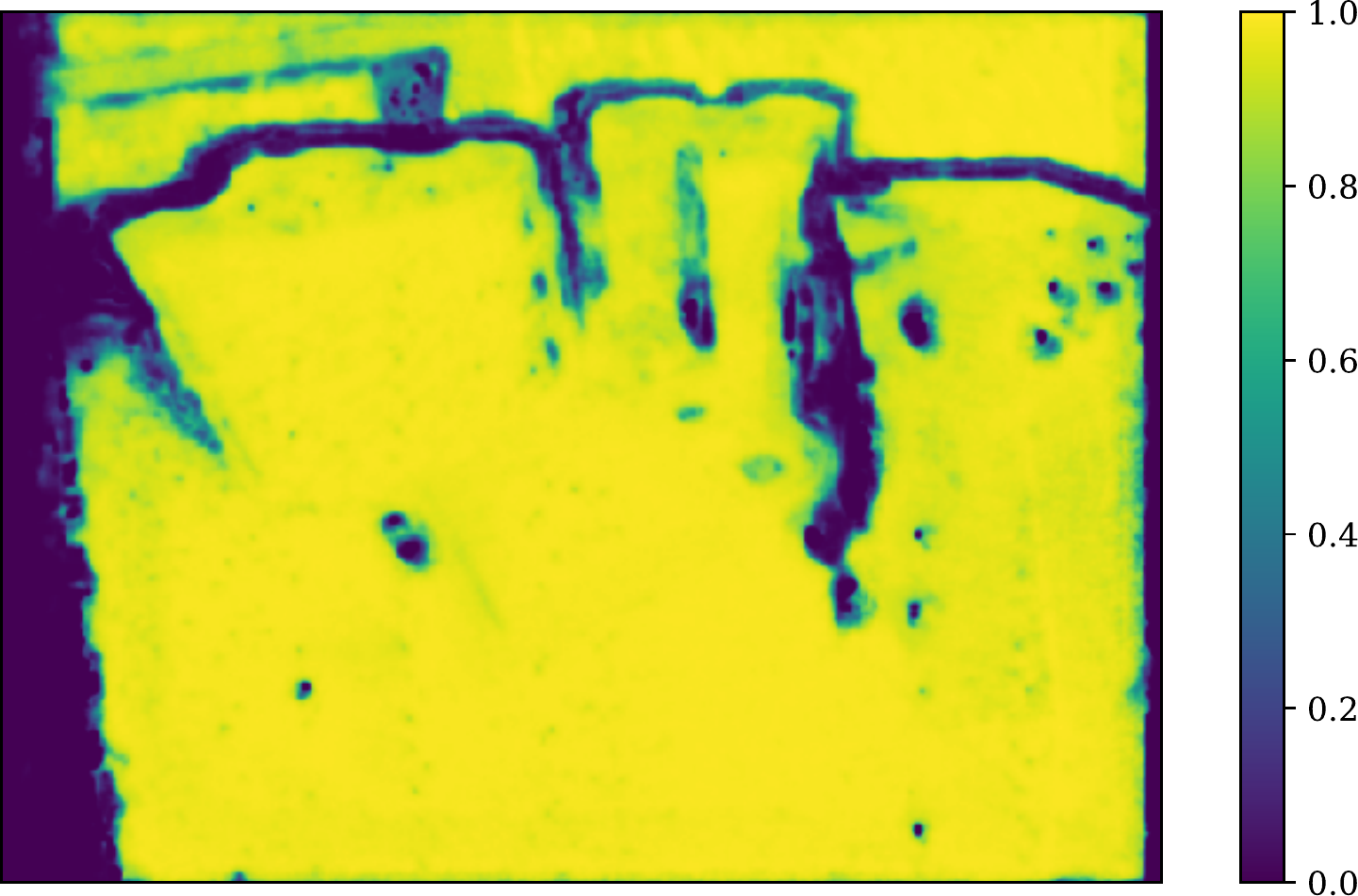}
    \caption{Confidence map.}
    \label{fig:resultsshapes:pinl}
    \end{subfigure}
    
    \caption{Results on a frame from the shapes dataset with (\subref{fig:resultsshapes:gt}) the ground truth, (\subref{fig:resultsshapes:asnselfsup}) original self-supervised ActiveStereoNet, (\subref{fig:resultsshapes:asnfinetuned}) ActiveStereoNet finetuned on \ourdatasetname{}, (\subref{fig:resultsshapes:ourwithoutliers}) ours with the outlier prediction module, (\subref{fig:resultsshapes:ourwithoutoutliers}) ours without the outlier prediction module,  and  (\subref{fig:resultsshapes:pinl}) the confidence map estimated by our module (validated pixels are those with a score equal or above $95\%$).}
    \label{fig:resultsshapes}
\end{figure*}

The most important benefit of our approach is the fact that the refinement module co-evolves with the outlier detection module. It enables the networks to focus, during training, on the pixels which it decides should be validated in the final semi-dense disparity map. As a result, the information the network is storing during the training phase is used more efficiently thus the performances on validated pixels should increase compared to a similar network which is not trained jointly with the outlier detection module. To measure this contribution, we trained a second version of the module without the outlier detection branch. We refer to this as the baseline module. We used a standard deviation of $1px$ for all pixels, with all other parameters such as the number of training epochs and the learning rate remaining the same. We then compared the baseline module with the full module (with outlier detection). As shown in Figure~\ref{fig:results:meanabserror}, the Bayesian model leads to significant improvements. As shown in Table~\ref{table:ourResults}, on average, the accuracy of the model with the outlier module is three times higher than the accuracy of the baseline model. This demonstrates that the proposed joint learning approach allows our network to greatly improve its performance on validated pixels by not wasting information on the pixels that have been invalidated by the outlier detection module. This is possible only because outliers are discarded. As shown in Figure~\ref{fig:prop_inliers}, the proportion of inliers varies significantly depending on the images. Some images can have less than 10\% inliers pixels, while others will reach around 90\%, with an average of 52\% of inliers for the \ourdatasetname{} test set; see Figure~\ref{fig:prop_inliers}. As shown in Table~\ref{table:ourResultsShapes}, our model with the outlier module is, relatively to the same model without the outlier module, four times as accurate, when evaluated on the Shapes dataset. This is globally coherent with the relative improvements with the results for the \ourdatasetname{} dataset.

\subsection{Comparison with state-of-the-art}
\label{sec:evaluation:comp}

\begin{table}[bt]
\centering
\begin{tabular}{|l|r|r|}
\hline
Method & \multicolumn{2}{c|}{MAE}\\
\hline
 & Validated pixels & {\color[HTML]{9B9B9B} All pixels} \\ \hline
 
 \multicolumn{3}{|c|}{Basline closed form solutions:} \\ \hline
Cost Interpolation \cite{jospin2021generalized} & 1.40px & {\color[HTML]{9B9B9B}7.67px} \\
Image Interpolation \cite{jospin2021generalized} & 1.39px & {\color[HTML]{9B9B9B}7.64px} \\ \hline

\multicolumn{3}{|c|}{Active Stereo models:} \\ \hline
ActiveStereoNet \cite{Zhang_2018_ECCV} & 1.30px & {\color[HTML]{9B9B9B}2.32px} \\
ActiveStereoNet \cite{Zhang_2018_ECCV} (finetuned) & 0.90px & {\color[HTML]{9B9B9B}1.82px} \\ \hline

\multicolumn{3}{|c|}{State-of-the-art large scale models:} \\ \hline
CascadeStereo \cite{gu2019cas} (finetuned) & 0.18px & {\color[HTML]{9B9B9B}\textbf{0.66px}} \\
ACVNet \cite{xu2022attention} (finetuned) & 0.22px & {\color[HTML]{9B9B9B} \textbf{0.66px}} \\ \hline

\multicolumn{3}{|c|}{Real time models:} \\ \hline
AnyNet \cite{AnyNet2019} (finetuned) & 0.53px & {\color[HTML]{9B9B9B} 2.06px} \\
RealTimeStereo \cite{Chang_2020_ACCV} (finetuned) & 0.74px & {\color[HTML]{9B9B9B} 1.64px} \\
MobileStereoNet \cite{shamsafar2022mobilestereonet} (finetuned) & 0.29px & {\color[HTML]{9B9B9B} 0.80px} \\ \hline

\multicolumn{3}{|c|}{Our model:} \\ \hline
Ours (w/o outlier module) & 0.36px & {\color[HTML]{9B9B9B}1.97px} \\
Ours (w/ outlier module) & \textbf{0.12px} & {\color[HTML]{9B9B9B}2.04px} \\ \hline

\end{tabular}
    \caption{Sub-pixel accuracy comparison in terms of MAE on the \ourdatasetname{} \cite{ourDataset} test set.}
    \label{table:ourResults}
\end{table}

\begin{table}[bt]
\centering
\scalebox{1.0}{
\begin{tabular}{|l|r|r|}
\hline
Method & \multicolumn{2}{c|}{MAE}\\
\hline
 & ~Validated pixels~ & {\color[HTML]{9B9B9B}~All pixels~} \\ \hline
 
 \multicolumn{3}{|c|}{Basline closed form solutions:} \\ \hline
Cost Interpolation \cite{jospin2021generalized}~ & 0.35px & {\color[HTML]{9B9B9B}7.29px} \\
Image Interpolation \cite{jospin2021generalized} & 0.33px & {\color[HTML]{9B9B9B}7.29px} \\ \hline

\multicolumn{3}{|c|}{Active Stereo models:} \\ \hline
ActiveStereoNet \cite{Zhang_2018_ECCV} & 0.89px & {\color[HTML]{9B9B9B}2.84px}\\
ActiveStereoNet (finetuned) & 0.81px & {\color[HTML]{9B9B9B}1.07px} \\ \hline

\multicolumn{3}{|c|}{State-of-the-art large scale models:} \\ \hline
CascadeStereo \cite{gu2019cas} (finetuned) & 0.61px & {\color[HTML]{9B9B9B}0.69px} \\
ACVNet \cite{xu2022attention} (finetuned) & 0.35px & {\color[HTML]{9B9B9B} \textbf{0.63px}} \\ \hline

\multicolumn{3}{|c|}{Real time models:} \\ \hline
AnyNet \cite{AnyNet2019} (finetuned) & 1.58px & {\color[HTML]{9B9B9B} 2.29px} \\
RealTimeStereo \cite{Chang_2020_ACCV} (finetuned) & 0.84px & {\color[HTML]{9B9B9B} 1.38px} \\
MobileStereoNet \cite{shamsafar2022mobilestereonet} (finetuned) & 0.55px & {\color[HTML]{9B9B9B} 0.79px} \\ \hline

\multicolumn{3}{|c|}{Our model:} \\ \hline
Ours (w/o outlier module) & 1.21px & {\color[HTML]{9B9B9B}5.40px} \\
Ours (w/ outlier module) & \textbf{0.32px} & {\color[HTML]{9B9B9B}6.17px} \\ \hline
\end{tabular}}
    \caption{Sub-pixel accuracy, in terms of MAE, of existing methods and our method on the Shapes test set.}
    \label{table:ourResultsShapes}
\end{table}

We compare the proposed model to the state-of-the-art active stereo model ActiveStereoNet \cite{Zhang_2018_ECCV}, using the pretrained weights provided by the authors~\cite{ActiveStereoNetCode}. However, since ActiveStereoNet was trained in a self-supervised fashion, we also finetuned it on the training dataset to obtain a more accurate and fair comparison. To this end, we used the training method described for StereoNet \cite{Khamis_2018_ECCV}. The Adaptive Robust Loss of~\cite{Barron_2019_CVPR} was used with the same meta-parameters as the original StereoNet \cite{Khamis_2018_ECCV}. We also finetuned a collection of real-time and state-of-the-art passive stereo methods to offer a larger and more recent base of comparison. We selected methods that have their codes and pretrained models publicly available. We used ACVNet \cite{xu2022attention}, CascadeStereo \cite{gu2019cas}, AnyNet \cite{AnyNet2019}, MobileStereoNet \cite{shamsafar2022mobilestereonet} and RealTimeStereo \cite{Chang_2020_ACCV}. ACVNet \cite{xu2022attention} was the top performing method on Kitti with both code and pretrained models available at the time of writing. CascadeStereo \cite{gu2019cas} was a model we experimented on and achieved satisfactory results for active stereo with while AnyNet \cite{AnyNet2019}, MobileStereoNet \cite{shamsafar2022mobilestereonet} and RealTimeStereo \cite{Chang_2020_ACCV} are real-time methods.

When all pixels are considered, ActiveStereoNet outperforms our method by a margin of around $0.22px (10\%)$; see Table~\ref{table:ourResults}. However, our model achieves significantly more accurate results, with an improvement of $0.78px (87\%)$, when the outlier detection module is included during training and only validated inliers are considered. Our model is even able to outperform large scale state-of-the-art models like CascadeStereo \cite{gu2019cas} and ACVNet \cite{xu2022attention}. Finetuning ActiveStereoNet improved the accuracy of pixels with large errors
by $0.5px (22\%)$ while pixels with the potential to be subpixel accurate were only improved by $0.4px (31\%)$, despite being trained with a robust loss. This shows that much of the training efforts of dense models like ActiveStereoNet are focused on high error pixels for which the model is unlikely to predict a highly accurate disparity in the first place. Most of the information stored by the network during training is not relevant in the case of semi-dense stereo. This is despite the fact that ActiveStereoNet was finetuned with a robust loss, which in itself should already help the network focus more on inliers. This highlights the benefits of our proposed Bayesian learning strategy for semi-dense models. It can achieve an accuracy of $0.12px$ and $0.32px$ for validated pixels; see Tables~\ref{table:ourResults} and \ref{table:ourResultsShapes} compared to $0.9px$ and $0.81px$ for ActiveStereoNet.

Note that the accuracy of our method on the Shapes dataset is not as good as its performance on the \ourdatasetname{} dataset. One possible explanation is that it is caused by small errors during the ground truth acquisition process for the Shapes dataset (see the Supplementary Material for more information). At the level of accuracy reached by our method, those small errors would become visible in the error metric. Figure~\ref{fig:resultsshapes} illustrates the key difference between how our semi-dense approach treats the disparities and how large-scale end-to-end models do. Our estimated depth map has visible artifacts (Fig.\ref{fig:resultsshapes:ourwithoutliers}) that the model just decides to ignore (Fig.\ref{fig:resultsshapes:pinl}) because it cannot be precise enough in those regions (Fig.\ref{fig:resultsshapes:ourwithoutoutliers}). On the other hand, ActiveStereoNet depth maps do not have artifacts (Fig.\ref{fig:resultsshapes:asnfinetuned}) but are over smoothed which degrades the subpixel accuracy of the model, even when restricted to our inliers, which should have been easier to match.

\section{Conclusion}
\label{sec::conclusion}
This paper proposes a lightweight yet high-performance deep learning model for disparity map refinement for active stereo matching. We demonstrated how a Bayesian model can be used to jointly improve the accuracy of depth predictions while efficiently detecting and invalidating outliers. The proposed architecture produces very accurate disparities on validated pixels, well above the state-of-the-art methods. Despite ignoring discarded pixels, our model performances are still competitive with the state-of-the-art dense methods on those pixels. This shows that state-of-the-art methods require significantly higher computational resources to deal with only a few outliers. Additionally, this work demonstrates the benefits of architectures tailored specifically to refine active stereo vision. Currently available end-to-end architectures require large amounts of computation and many of their functions are redundant with traditional matching algorithms in active stereo vision. Our model is also able to generalize to passive stereo, reaching an accuracy on par with the
state-of-the-art semi-dense passive stereo models, albeit at
the cost of a drastic reduction in validated pixels density.


%





%


{\small
\bibliographystyle{IEEEtran}
\bibliography{references}
}

\begin{IEEEbiography}[{\includegraphics[width=1in,height=1.25in,clip,keepaspectratio]{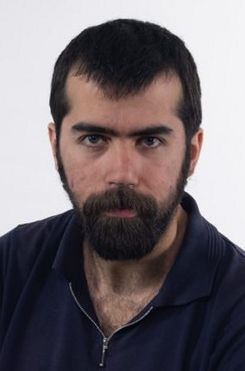}}]{Laurent Valentin Jospin}  received the MSc degree in environmental science in engineering from EPFL in 2017, with a minor in computational science. Since 2019 he is a Ph.D. research student in the field of deep learning for computer vision at the University of Western Australia. His main research interests include 3d reconstruction, sampling and image acquisition strategies, computer vision applied to robotic navigation, computer vision applied to environmental sciences, and Bayesian statistics applied to computer vision.
\end{IEEEbiography}

\begin{IEEEbiography}[{\includegraphics[width=1in,height=1.25in,clip,keepaspectratio]{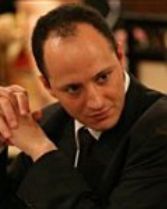}}]{Hamid Laga} received the MSc and PhD degrees in Computer Science from Tokyo Institute of Technology in 2003 and 2006, respectively. He is currently a Professor at Murdoch University (Australia). His research interests span various fields of machine learning, computer vision, computer graphics, and pattern recognition, with a special focus on the 3D reconstruction, modeling, and analysis of static and deformable 3D objects, and on image analysis and big data in agriculture and health. He is the recipient of the Best Paper Awards at SGP2017, DICTA2012, and SMI2006.
\end{IEEEbiography}

\begin{IEEEbiography}[{\includegraphics[width=1in,height=1.25in,clip,keepaspectratio]{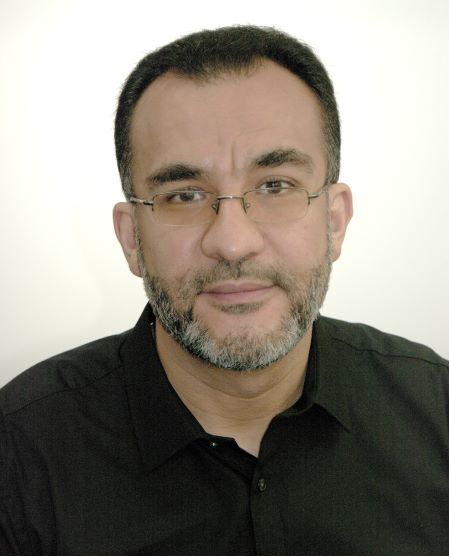}}]{Farid Boussaid} received the M.S. and Ph.D.
degrees in microelectronics from the National Institute of Applied Science (INSA), Toulouse, France, in 1996 and 1999 respectively. He joined Edith Cowan University, Perth, Australia, as a Postdoctoral Research Fellow, and a Member of the Visual Information Processing Research Group in 2000. He joined the University of Western Australia, Crawley, Australia, in 2005, where he is currently a Professor. His current research interests include neuromorphic engineering, smart sensors, and machine learning.
\end{IEEEbiography}

\begin{IEEEbiography}[{\includegraphics[width=1in,height=1.25in,clip,keepaspectratio]{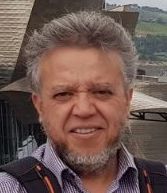}}]{Mohammed Bennamoun} is a Winthrop Professor
in the Department of Computer Science and Software Engineering at the University of Western Australia (UWA) and is a researcher in computer vision, machine/deep learning, robotics, and signal/speech processing. He has published 4 books (available on Amazon), 1 edited book, 1 Encyclopedia article, 14 book chapters, 180+
journal papers, 260+ conference publications, 16 invited and keynote publications. His h-index is 65 and his number of citations is 18,200+ (Google Scholar). He was awarded 70+ competitive research grants, from the Australian Research Council, and numerous other Government, UWA, and industry Research Grants. He successfully supervised +26 Ph.D. students to completion. He won the Best Supervisor of the Year Award at Queensland University of Technology (1998) and received the award for research supervision at UWA (2008 and 2016) and Vice-Chancellor Award for mentorship (2016). He delivered conference tutorials at major conferences, including IEEE CVPR 2016, Interspeech 2014, IEEE ICASSP, and ECCV. He was also invited to give a Tutorial at an International Summer School on Deep Learning (DeepLearn 2017).
\end{IEEEbiography}

\end{document}


\title{Bayesian Learning for Disparity Map Refinement for Semi-Dense Active Stereo Vision -- Supplementary material}

\author{Laurent Valentin Jospin,
        Hamid Laga,
        Farid Boussaid,
        and~Mohammed Bennamoun,~\IEEEmembership{Senior Member,~IEEE}
\thanks{L.V. Jospin, F. Boussaid and M. Bennamoun are with the University of Western Australia.}
\thanks{H. Laga is with Murdoch University and the University of South Australia.}}

\maketitle

\section{Theoretical guarantee on model performance}
\label{appendix:improvementdemo}

Let  $f : I \rightarrow O$ be a function where:
\begin{itemize}
    \item $I$ is a given input space, in our case a patch of pixels in a raw disparity map, along with a slice of the corresponding cost volume and support indicator function, and 
    \item $O$ is an output space, in our case all possible refined disparities. 
\end{itemize}
\noindent Let also $A : I \rightarrow O$ be a set of functions with the same input and output space, in our case all functions encoded by the considered neural network architecture. 

The aim of the learning process is to find $a \in A$ such that  for a given distance $d_f : (I,O) \rightarrow \mathbb{R}^{+}$   to the function $f$ is such that 
\begin{equation}
    \int_I d_{f}(x, a(x)) dx  \le \int_I d_{f}(x, a'(x)) dx  ~ \forall a' \in A.
\end{equation}

\noindent Now, given a set of functions $\Lambda : I \rightarrow \{0,1\}$ with $\lambda(I) = 1 \in \Lambda$, which represents possible classifications of the inputs as outliers, and a scale factor $\alpha \in (0,1)$, the aim of learning to approximate $f$ for a given outlier classification $\lambda$ is to find $s \in A$ such that

\begin{equation}
    d(s, \lambda) \le d(a', \lambda)  ~ \forall a' \in A,
\end{equation}
\noindent with 
\begin{equation}
    d(a, \lambda) = \int_I (1 + (\alpha - 1) \lambda(x)) d_{f}(x, a(x)) dx.
\end{equation}

\noindent The key observation here is that $s$ is always as close or closer to $f$ as $a$ when the proximity is measured only on inliers. To demonstrate this,  let's proceed by contradiction and assume that we have \begin{equation}
    \int_I (1 - \lambda(x)) \cdot (d_{f}(x, a(x)) - d_{f}(x, s(x))) dx < 0.
\end{equation}

\noindent To keep the definition of $s$ valid it means we have:

\begin{equation}
\begin{aligned}[c]
    \int_I \alpha \lambda(x) \cdot (d_{f}(x, a(x)) - d_{f}(x, s(x))) dx > \\
    \int_I (1 - \lambda(x)) \cdot (d_{f}(x, s(x)) - d_{f}(x, a(x))) dx.
\end{aligned}
\end{equation}

\noindent But this implies, by linearity of the integral, that:

\begin{equation}
\begin{aligned}[c]
   \int_I \lambda(x) \cdot (d_{f}(x, a(x)) - d_{f}(x, s(x))) dx > \\
   \int_I (1 - \lambda(x)) \cdot (d_{f}(x, s(x)) - d_{f}(x, a(x))) dx,
\end{aligned}
\end{equation}

\noindent which is in contradiction with the definition of $a$. Thus, $s$ performs at least as good as $a$ for the inliers. This demonstration generalizes for stochastic models where $\lambda$ is not a binary value but a probability for a given point to be an outlier.

\section{Adaptive Training Strategy}
\label{sec:training}

\begin{figure*}[tb]
    \centering
    \begin{subfigure}[t]{0.48\textwidth}
    \includegraphics[width=\textwidth]{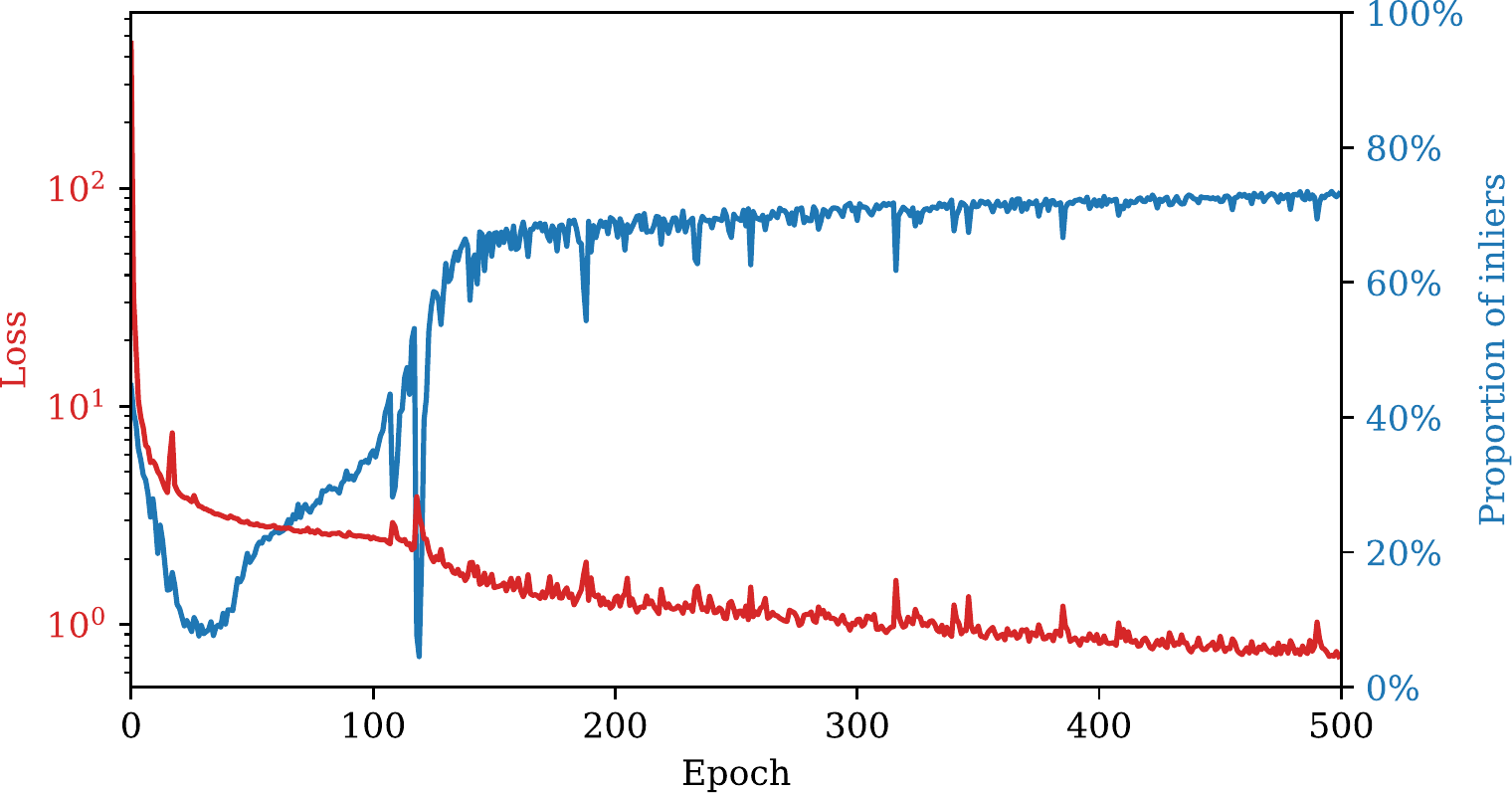}
    \caption{Successful training on \ourdatasetname{} without adaptive training}
    \label{fig:train_strat:without_adaptive_sucess}
    \end{subfigure}
    \hfill
    \begin{subfigure}[t]{0.48\textwidth}
    \includegraphics[width=\textwidth]{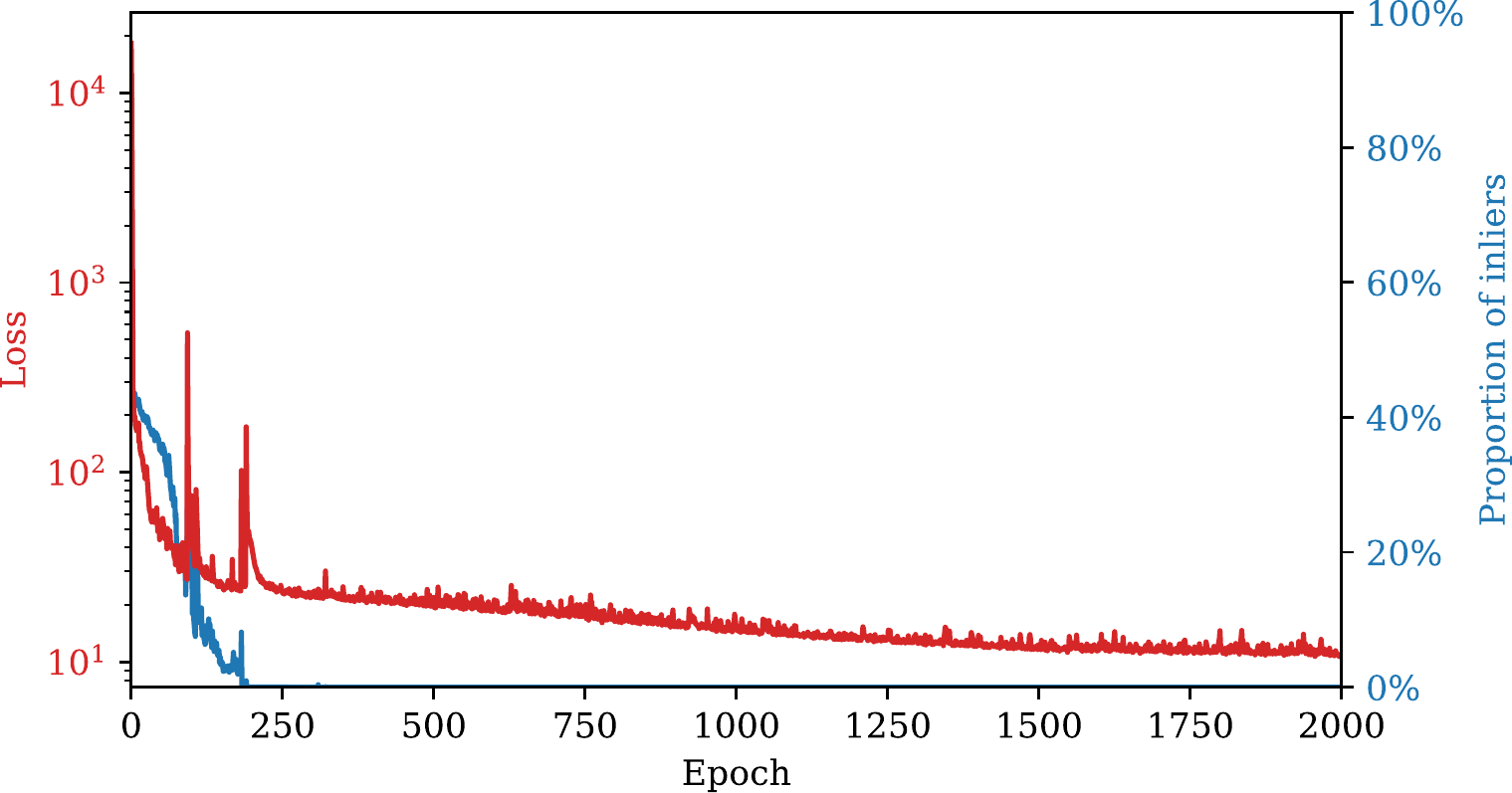}
    \caption{Failed training on the Middleburry dataset without adaptive training.}
    \label{fig:train_strat:without_adaptive_failure}
    \end{subfigure}
    
    \begin{subfigure}[t]{0.48\textwidth}
    \includegraphics[width=\textwidth]{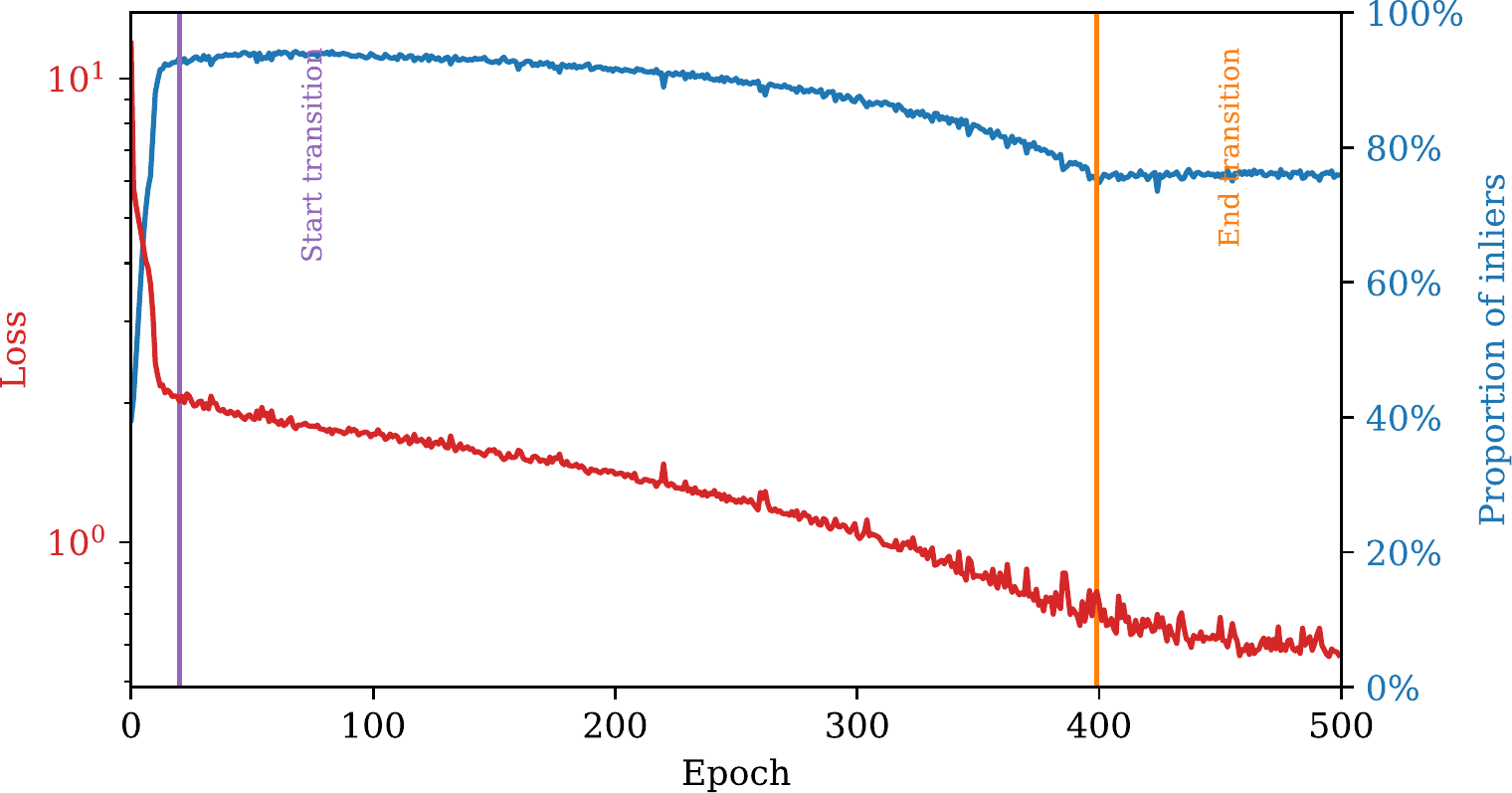}
    \caption{Successful training on \ourdatasetname{} with adaptive training.}
    \label{fig:train_strat_on:apstereo}
    \end{subfigure}
    \hfill
    \begin{subfigure}[t]{0.48\textwidth}
    \includegraphics[width=\textwidth]{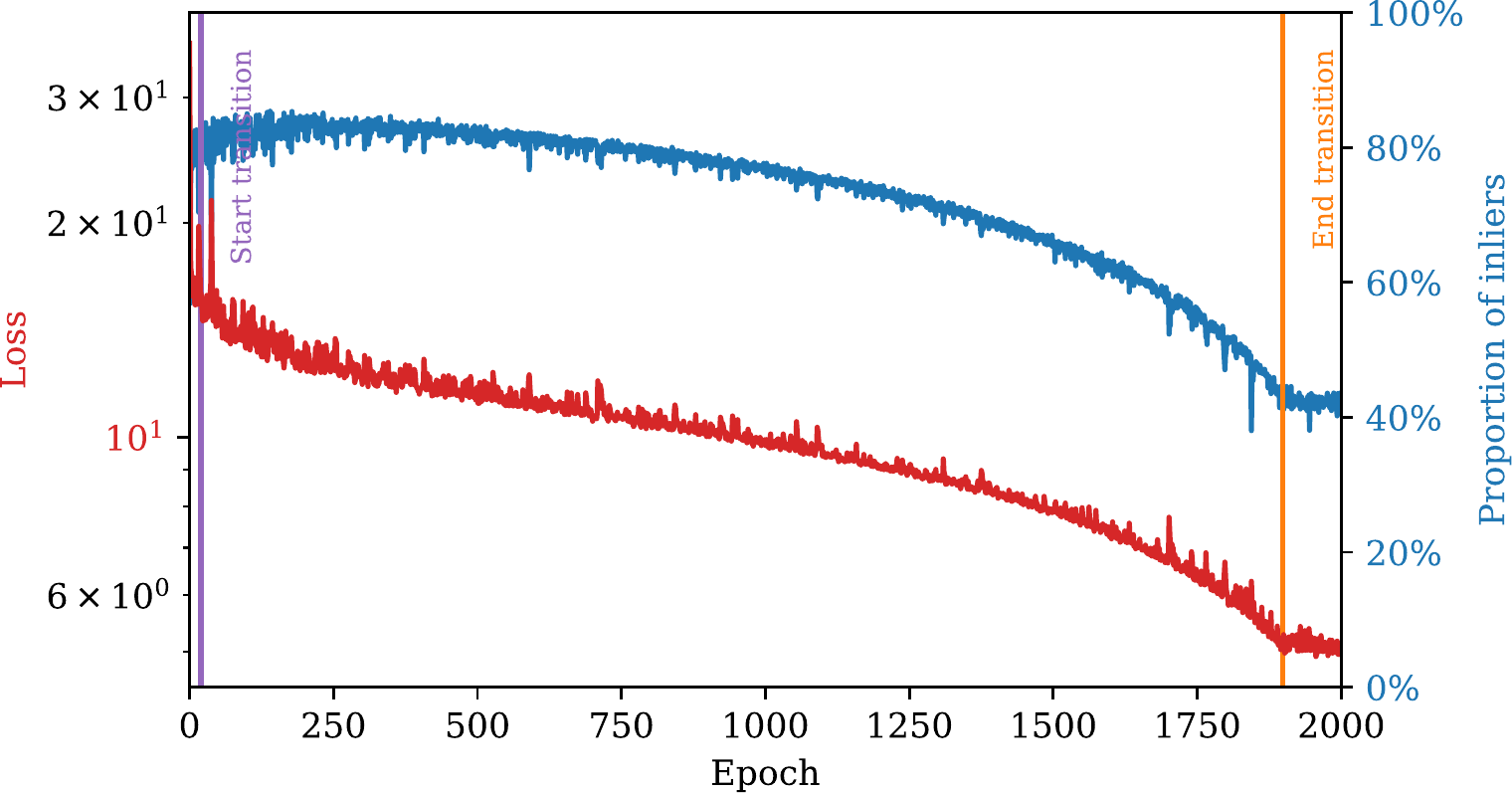}
    \caption{Successful training on Middleburry with adaptive training.}
    \label{fig:train_strat_on:middleburry}
    \end{subfigure}
    
    \caption{Evolution of the loss and average proportion of inliers without and with using an adaptive training scheme. The proportion of inliers is estimated by taking the average over all pixels of $(1-p_{out})$.}
    \label{fig:train_strat}
\end{figure*}

The challenge with the proposed approach is that even if the optimum of the loss function has very good properties, \ie it encourages the network to distinguish between inliers and outliers based solely on the expected accuracy and requires no other hypothesis, finding such optimum is hard. Gradient descent algorithms tend to promote outliers predictions during their early steps, limiting the ability of the regression module and the outlier classification module to co-evolve. Worse, this will also slow the convergence towards a good set of parameters, leaving  the loss to plateau in a region where the network classifies all points as outliers. For active stereo datasets with a high proportion of inliers, like \ourdatasetname{}, this is a moderate problem; see Fig.~\ref{fig:train_strat:without_adaptive_sucess}. However,  for certain starting configurations, especially if the dataset is more challenging like, \eg the Middleburry passive stereo dataset \cite{10.1007/978-3-319-11752-2_3}, the optimization can get stuck in a local optima and always predict all pixels as outliers; see Figure~\ref{fig:train_strat:without_adaptive_failure}.

To address this issue, we experimented with a training scheme, hereinafter referred to as adaptive scheme,  where the stochastic model is adaptive. More specifically, we propose to set the standard error for inliers and outliers to a value $\sigma_0$ at the start of the training. This will have the effect of training the regression module while the classification module is kept idle. We keep those standard error constants for an initial burn-in number of epochs before continuing with a transition period, where the paramaters of the stochastic model smoothly transition toward their values, and finally a certain number of epochs where the standard error in the stochastic model keep their final values. In more details, the probabilities of inliers or outliers are computed at a given epoch $e$ as:
\begin{equation}
    \sigma(e) = \sigma_0 + \left( [e>e_t] \dfrac{\min(e_f,e)-e_t}{e_f-e_t} \right)  (\sigma_f - \sigma_0),
\end{equation}

\noindent where $e_t$ is the first epoch of the transition phase, $e_f$ is the first epoch after the transition phase, and $\sigma_f$ is the true (or final) standard error in the stochastic model for either the inliers or outliers.

Additionally, we add a penalty to the loss to encourage more pixels to be classified as inliers. The formulation is a reward for inliers, which converge towards $0$ as training progresses:
\begin{equation}
    \left( [e>e_t] \dfrac{\min(e_f,e)-e_t}{e_f-e_t} \right) \times \lambda \times \bar{p}_{out},
\end{equation}

\noindent where $\lambda$ is a constant set by the operator, and $\bar{p}_{out}$ is the mean probability of any given pixel in the current minibatch being an outlier.

The results (Fig.~\ref{fig:train_strat}) show that, during the initial training phase, the proportion of inliers is much larger, allowing for more co-evolution between the outliers detection module and the disparity regression module. When compared to a successful training without the adaptive scheme (Fig.~\ref{fig:train_strat:without_adaptive_sucess}), the proposed training method leads to marginal improvements in terms of the proportion of validated pixels. 

Note that, while the proposed adaptive training strategy breaks a few assumptions of our model during training, especially by having the model validate more pixels instead of figuring out by itself what the actual proportion of inliers in the data is, the final training epochs still use only the Bayesian model described in Section 3.1 of the main paper. A corollary is that as long as the model has converged to a, possibly local, minima of the objective function with our adaptive scheme, there exist a certain starting configuration of the network and a  corresponding gradient descent scheme that will converge to the same (possibly local) minima of the objective function without the adaptive training scheme. This result means that, overall, our adaptive training scheme does not break our model hypothesis. It just makes it more likely for the model to converge to a configuration with more inliers. This is achieved by forcing the training algorithm to traverse regions of the parameters space where the network predicts a high proportion of inliers.

\section{The "Shapes" dataset}
\label{appendix:realimages}

\begin{figure*}[tb]
    \centering
    \begin{subfigure}[b]{0.28\textwidth}
    \centering
    \includegraphics[height=95pt]{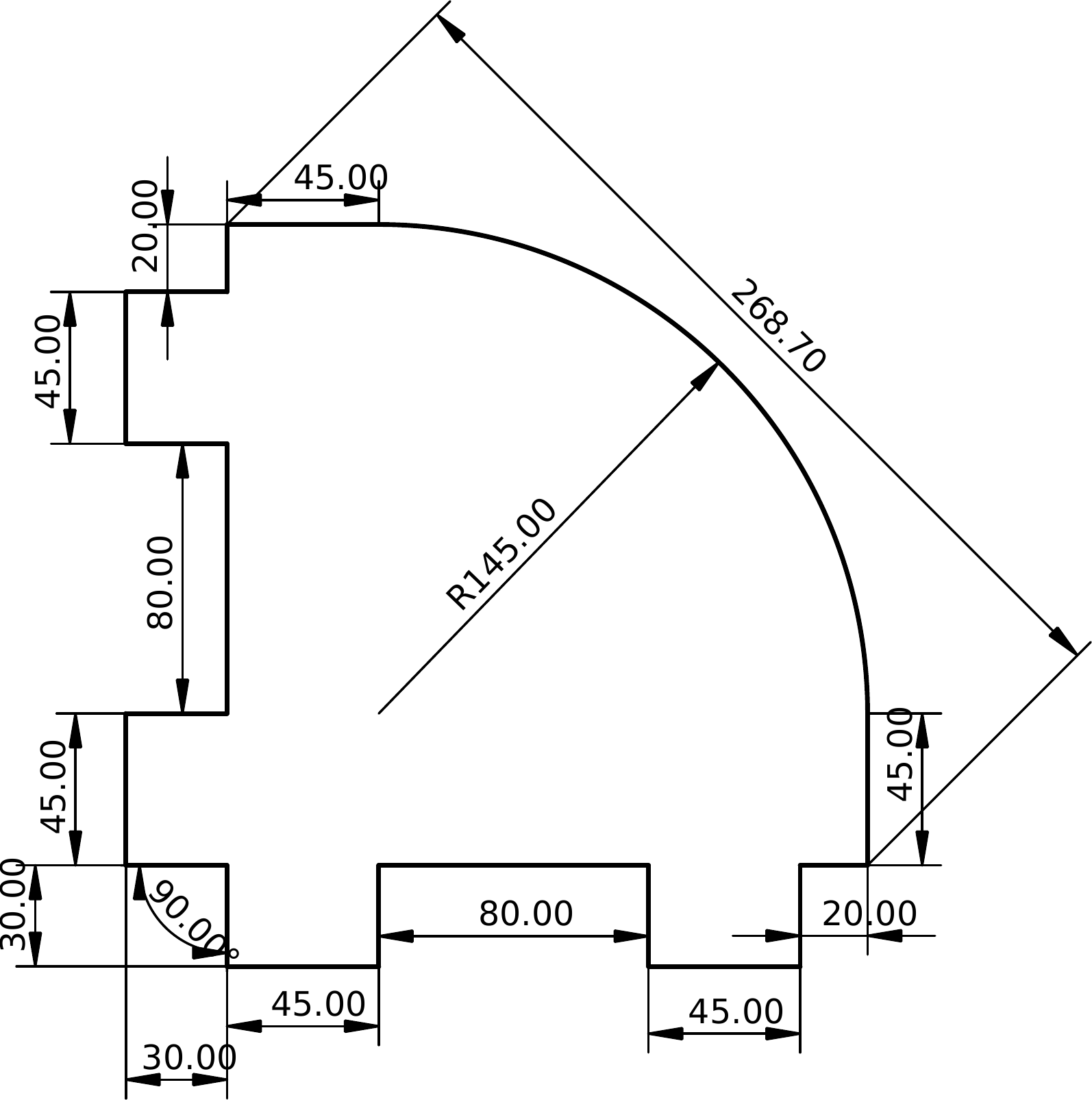}
    \caption{Shape1}
    \label{fig:shapes:one}
    \end{subfigure}
    \hfill
    \begin{subfigure}[b]{0.32\textwidth}
    \centering
    \includegraphics[height=90pt]{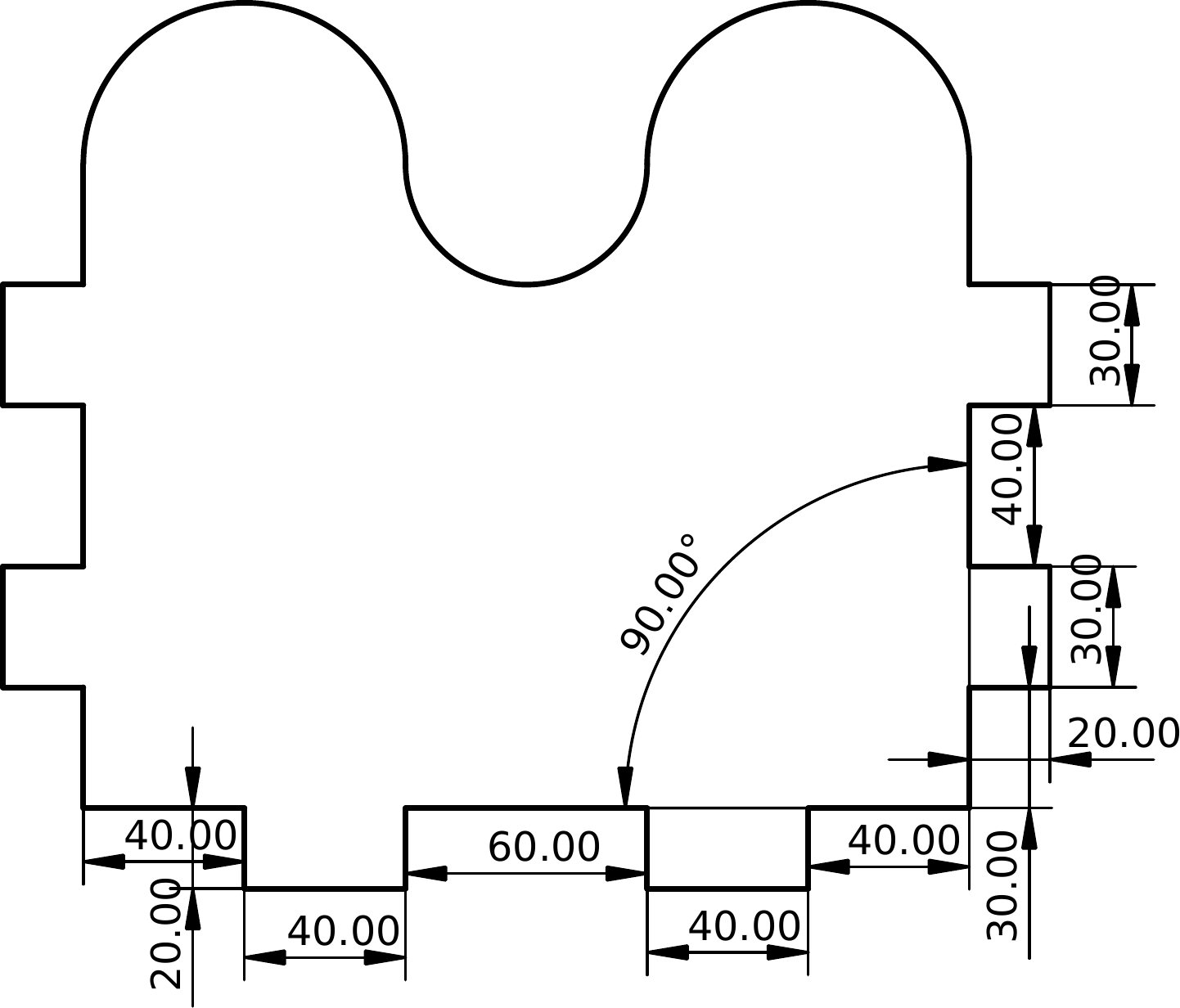}
    \caption{Shape2}
    \label{fig:shapes:two}
    \end{subfigure}
    \hfill
    \begin{subfigure}[b]{0.28\textwidth}
    \centering
    \includegraphics[height=90pt]{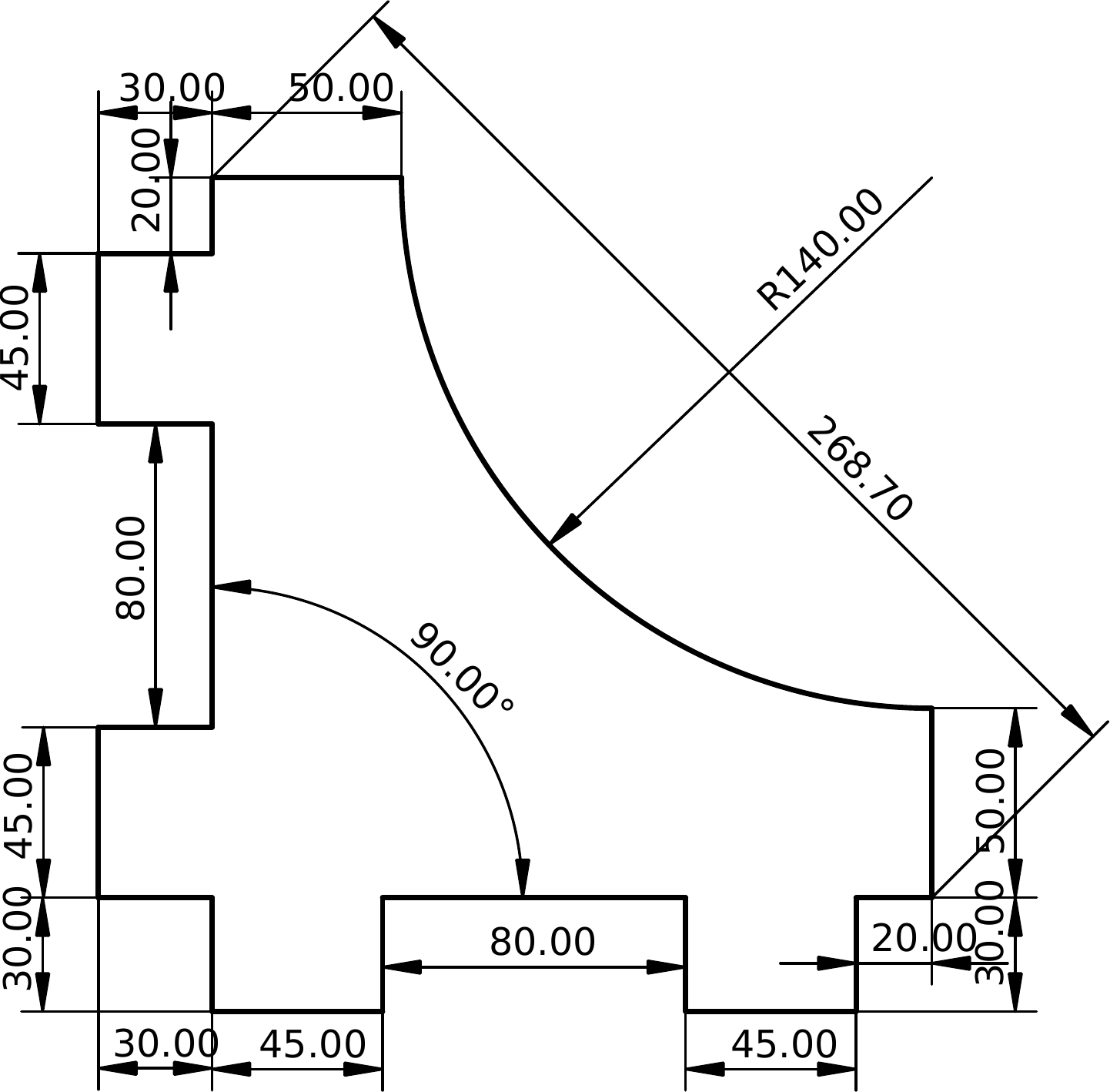}
    \caption{Shape3}
    \label{fig:shapes:three}
    \end{subfigure}

    \begin{subfigure}[b]{0.4\textwidth}
    \centering
    \includegraphics[height=85pt]{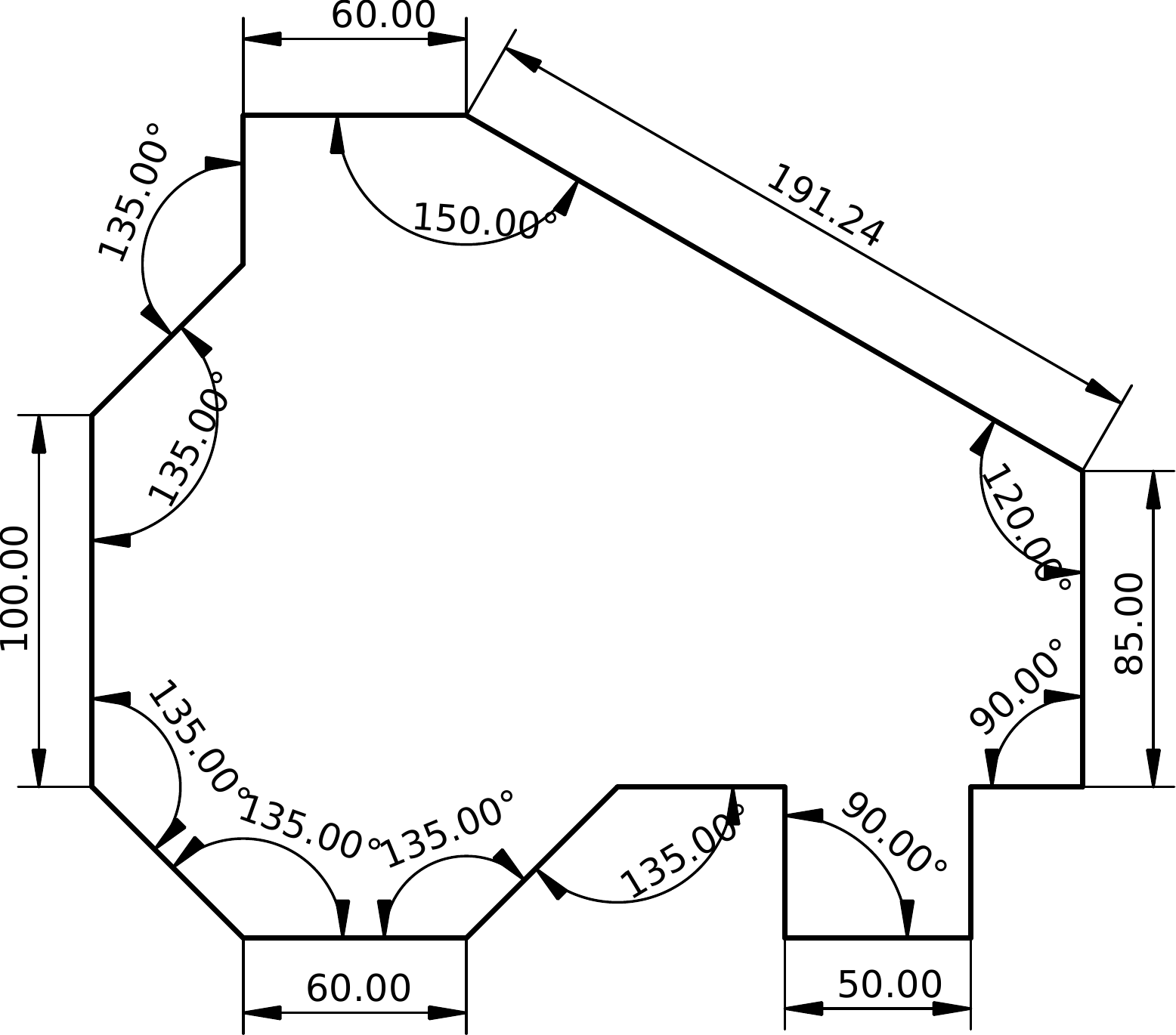}
    \caption{Shape4}
    \label{fig:shapes:four}
    \end{subfigure}
    \hspace{15pt}
    \begin{subfigure}[b]{0.4\textwidth}
    \centering
    \includegraphics[height=100pt]{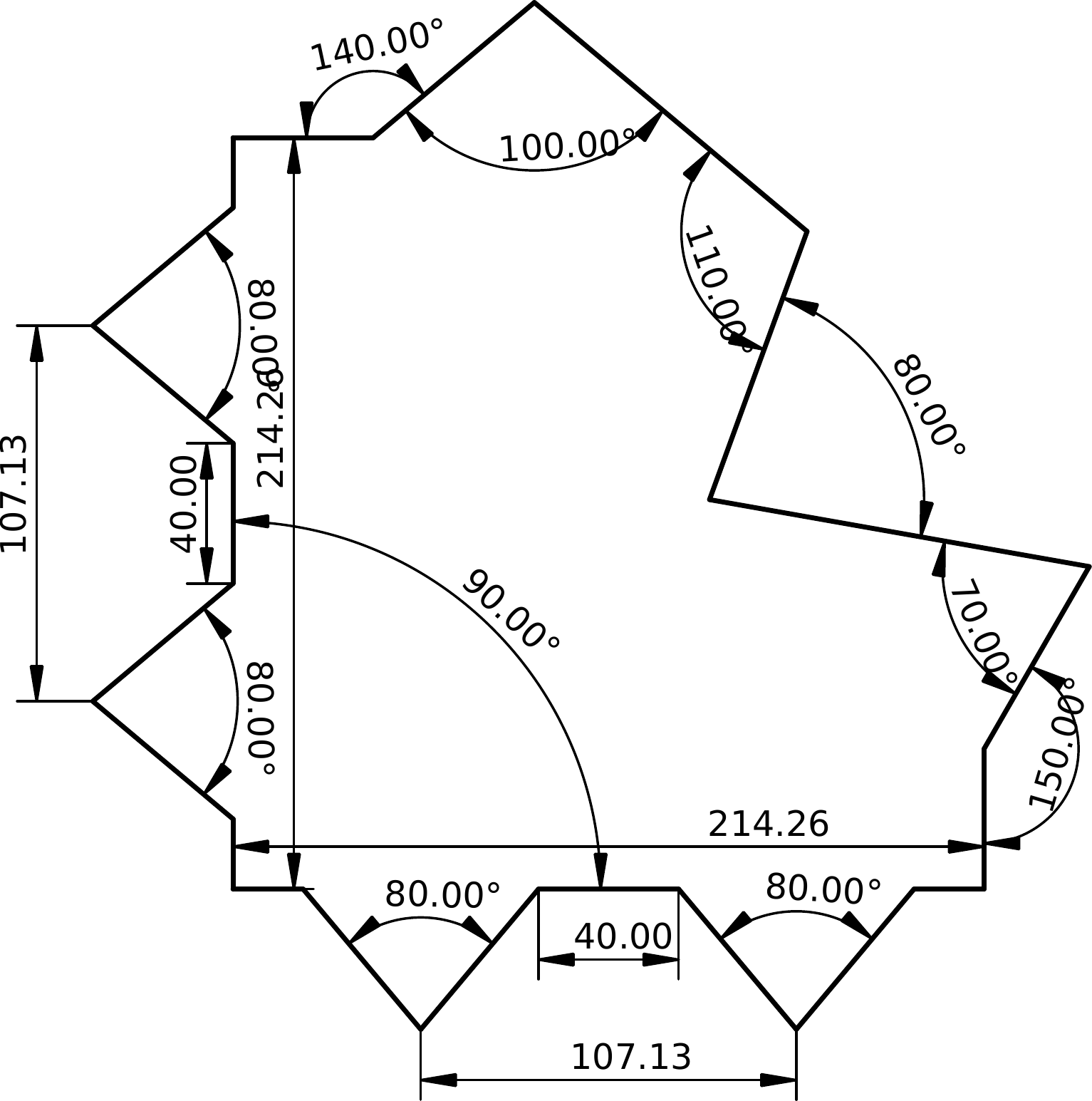}
    \caption{Shape5}
    \label{fig:shapes:five}
    \end{subfigure}
    
    \caption{The shapes used to build our dataset.}
    \label{fig:shapes}
\end{figure*}

The Shapes dataset is a dataset of active stereo images pairs acquired with a D435i RealSense camera. To get a highly accurate 3D ground truth, we first used a computer-aided design (CAD) software to design a series of 3D objects with a variety of shapes and angles (Fig.~\ref{fig:shapes}). Each 3D object exhibits a series of corners that can be easily identified in the images. We then mandated a specialized workshop to cut these shapes into sheets of $200$mm thick high-grade foamed polystyrene to produce volumetric shapes with known dimensions, up to mm precision; see Fig.~\ref{fig:shapes}. 

The shapes were then arranged to form static scenes, which we captured from multiple directions using a D435i Realsense camera. We used the maximum resolution of the camera, \ie $1280\times800px$, and a bit depth of $16$ bits per channel. This gives the maximum information available to any method benchmarked on the data.

The corners of the different shapes were then manually labelled using a photogrammetry toolbox implemented using g2o \cite{g2o}. To ensure the reconstruction is as accurate as possible, angle and length constraints were entered into the model, measured from the CAD models of the shapes. After the bundle adjustment and the rectification of the stereo images, the CAD models were aligned onto the images to generate the ground truth disparities.

Based on the estimated a-posteriori accuracy of our fitting model, we found that the positions of the shapes are known with a precision of around $1$mm. With the additional uncertainty due to the maximum precision of the cutting process, a standard deviation of at least $\sqrt{2}$mm is to be expected. This translates to a disparity error of $0.5$ pixels when $300$mm away from the camera and $0.2$ pixels when $500$mm away from the camera. Hence, while highly precise, the ground truth provided with the data has a small systematic error, which can affect the measurement performance for methods with subpixel accuracy well below $1px$ (like ours).

We downscaled and cropped the images to $640\times480$px in order to reduce the impact of those artifacts and to remove the blackened-out image regions which resulted from the rectification process. The scale factor used is $5/8$, which reduces the error in the disparity estimate by a similar factor, down to $0.31$ pixels when $300$mm away from the camera and $0.13$ pixels when $500$mm away from the camera.

The final dataset is made of $12$ images from $3$ different scenes, with disparities between roughly 20px and 100px.

\begin{table*}[t]
    \centering
    \begin{tabular}{|l|rrrrrrrrrr|}
    \hline
        \textbf{Method}	& \textbf{Adirondack} & \textbf{Jadeplant} & \textbf{Motorcycle} & \textbf{Piano} & \textbf{Pipes} & \textbf{Playroom} & \textbf{Playtable} & \textbf{Recycle} & \textbf{Shelves} & \textbf{Vintage} \\
	\hline		
DCNN \cite{semi-dense-stereo2019} & 0.33 & 0.31 & \textbf{0.33} & \textbf{0.40} & 0.30 & 0.49 & \textbf{0.25} & \textbf{0.37} & \textbf{0.43} & \textbf{0.39} \\
R-NCC \cite{LI20191318} & 0.4 & 0.55 & 0.57 & 0.5 & 0.48 & 0.66 & 0.39 & 0.49 & 0.69 & 0.45 \\
CVANet RVC  & 0.5 & 1.69 & 0.66 & 0.5 & 0.91 & 0.72 & 0.54 & 0.4 & 0.48 & 0.49 \\
RASNet & 0.38 & 3.23 & 0.53 & 0.46 & 0.74 & 0.68 & 0.37 & 0.46 & 0.62 & 0.9 \\
AANet RVC \cite{Xu_2020_CVPR} & 0.43 & 1.77 & 0.83 & 0.69 & 0.89 & 0.86 & 0.47 & 0.39 & 0.67 & 1.13 \\
SUWNet \cite{9191126} & 0.62 & 1.89 & 0.77 & 0.63 & 1.04 & 0.89 & 0.65 & 0.54 & 0.65 & 0.64 \\
AANet++ \cite{Xu_2020_CVPR} & 0.49 & 2.14 & 0.91 & 1.11 & 1.17 & 1.01 & 0.59 & 0.44 & 0.78 & 0.98 \\
DISCO \cite{8784869} & 0.64 & 2.3 & 0.83 & 0.88 & 1.17 & 1.13 & 0.64 & 0.53 & 0.88 & 1.48 \\
LEAStereo \cite{NEURIPS2020_fc146be0} & 0.63 & 2.66 & 0.89 & 1.09 & 1.34 & 1.21 & 0.78 & 0.66 & 0.98 & 1.21 \\
HITNet \cite{tankovich2021hitnet} & 0.42 & 2.73 & 0.79 & 1.14 & 1.11 & 1.21 & 0.59 & 0.51 & 0.91 & 0.72 \\
\hline
 \textbf{Ours} & \textbf{0.21} & \textbf{0.22} & 0.66 & 0.46 & \textbf{0.22} & \textbf{0.48} & 0.33 & 0.44 & 0.45 & 0.72 \\ 
\hline
    \end{tabular}
    \caption{Sub-pixel accuracy (MAE) of our method, trained on our simulated dataset only, compared to the top-ten semi-dense stereo methods in the Middleburry challenge. All scores are from the semis-dense Middlebury Stereo Evaluation Benchmarck - Version 3 \cite{MiddleburryBenchmark} and represent the mean absolute error of each method on its specific set of inliers. All errors are scaled to correspond to the full resolution dataset, even if each method is free to use either the full half or quarter resolution dataset.}
    \label{tab:middleburry_results}
\end{table*}

\begin{figure}[thb]
    \centering
    
    \begin{subfigure}[b]{0.45\textwidth}
    \includegraphics[width=\textwidth]{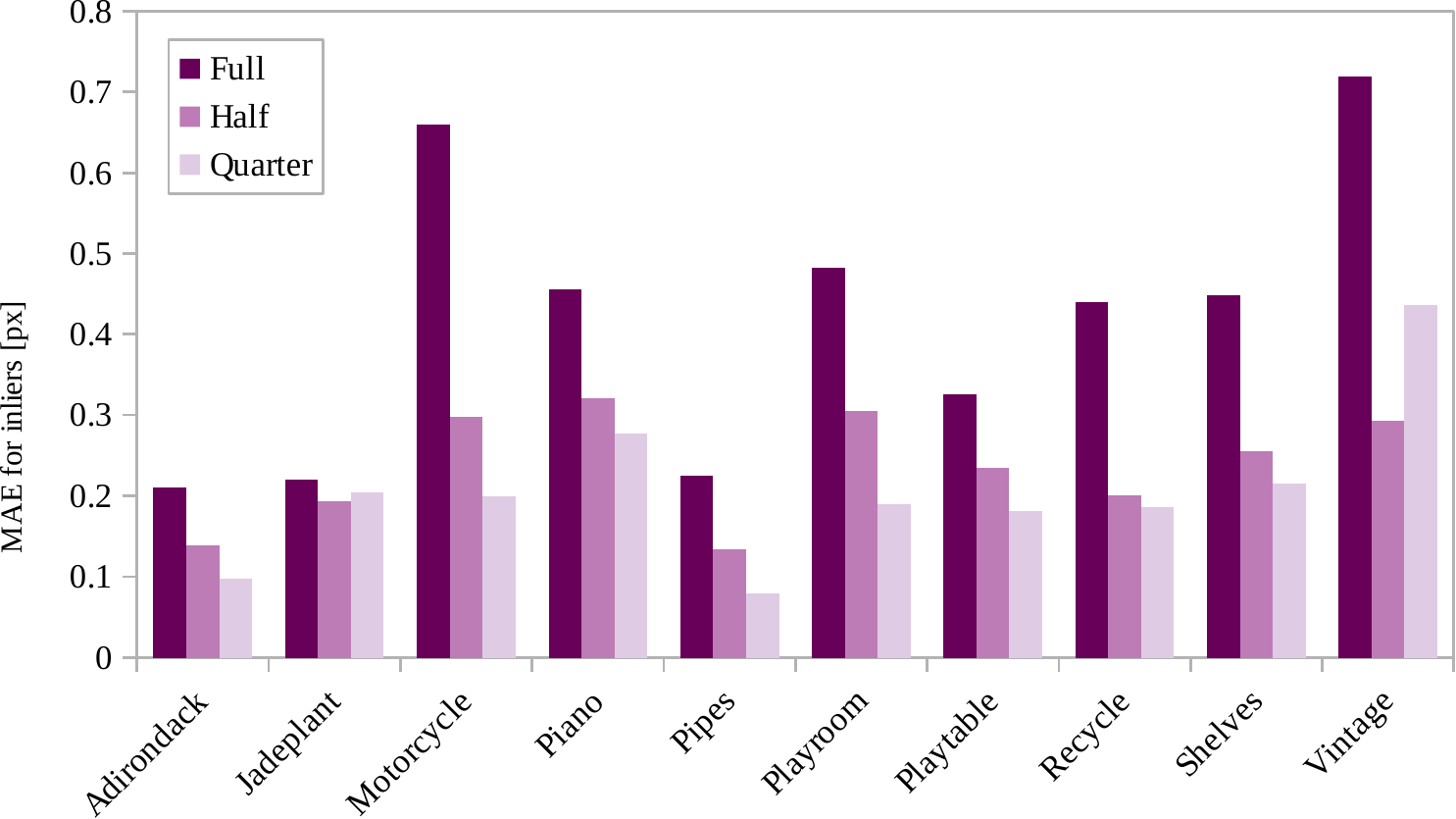}
    \caption{MAE computed on areas where $p_{out} \le 5\%$.}
    \label{fig:results_middleburry:meanabserror}
    \end{subfigure}
   
    \begin{subfigure}[b]{0.45\textwidth}
    \includegraphics[width=\textwidth]{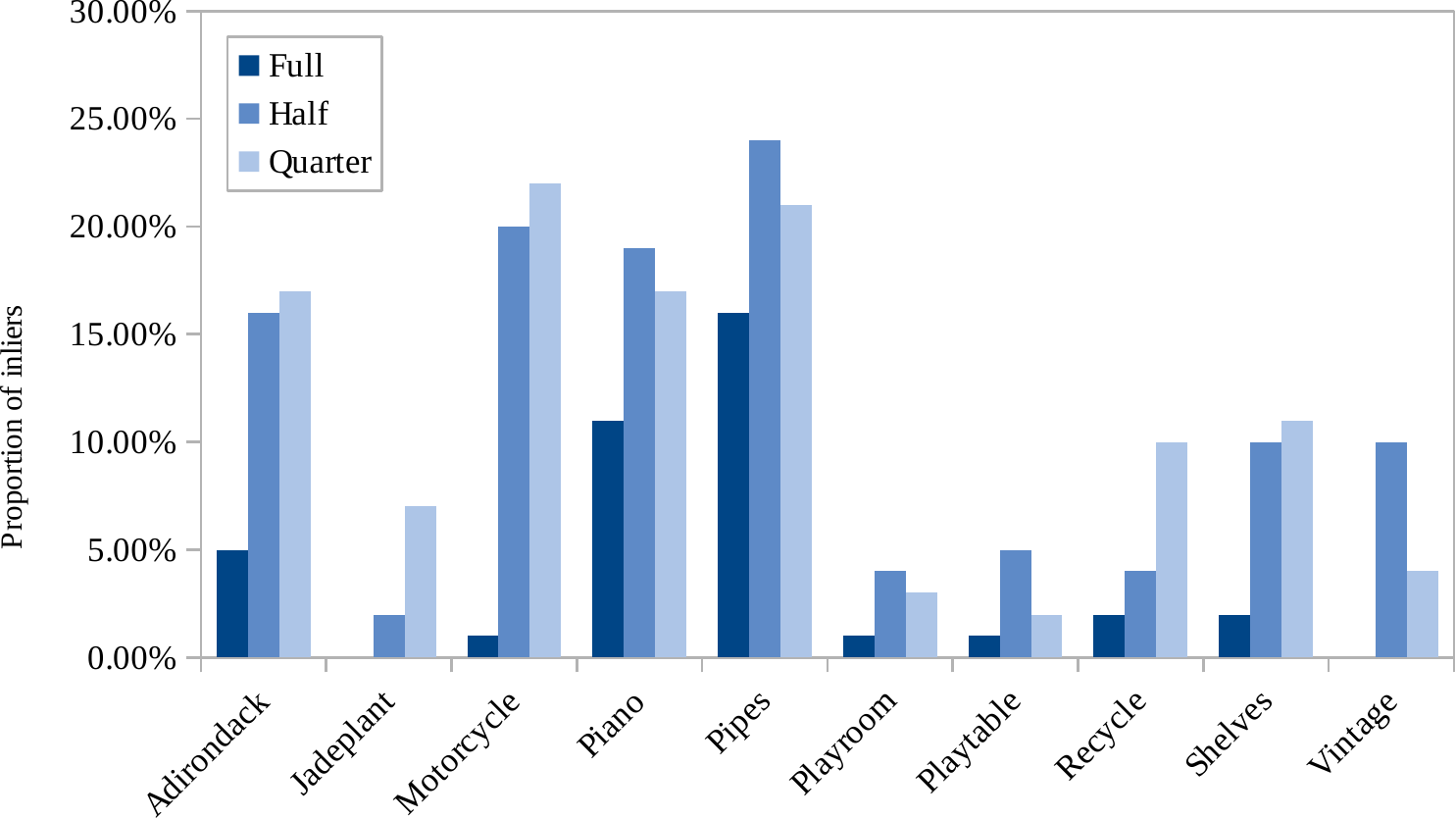}
    \vspace{-17pt}
    \caption{Proportion of pixels selected, \ie pixels where $p_{out} \le 5\%$.}
    \label{fig:results_middleburry:pixelsselected}
    \end{subfigure}
    
    
    \caption{MAE of the proposed model on on inliers for the Middleburry train set with all three possible resolutions.}
    
    \label{fig:results_middleburry}
\end{figure}

\subsection{Generalization abilities}
\label{sec:evaluation:comppassive}

In addition to our experiments on active stereo, we have tested the proposed model on the Middleburry 2014 stereo dataset \cite{10.1007/978-3-319-11752-2_3}. The evaluation benchmark \cite{MiddleburryBenchmark} provided with the dataset contains a semi-dense stereo vision section where each method provides a validated set of pixels, in addition to the disparity estimation. The final scores displayed on the website were calculated only from validated pixels. In our evaluation, we did not fine-tune our model using the training set from Middleburry 2014. In Table~\ref{tab:middleburry_results}, we compare the MAE accuracy of our method against the top 10 semi-dense matching methods on the ”train sparse” and ”avgerr” set category of the benchmark, which lists the MAE of semi-dense methods. Even without any fine-tuning, our method achieves performance comparable to the state-of-the-art, providing the top performance on 4 out of 10 images.

As can be seen in Figure~\ref{fig:results_middleburry:pixelsselected}), the proportion of pixels classified as inliers with our method is relatively low. However, this is to be expected since matching patches in passive stereo is more ambiguous than in active stereo. This means that a smaller proportion of pixels will be inliers. However, our model is not designed for passive stereo. Therefore, the fact that the proposed method is still able to reliably detect pixels when presented with data from a completely different domain shows its robustness.

{\small
\bibliographystyle{IEEEtran}
\bibliography{references}
}